\documentclass[twocolumn]{autart}

\usepackage{xcolor}
\usepackage{amsmath}
\usepackage{amssymb}
\usepackage{amsfonts}
\usepackage{mathtools}
\usepackage{graphicx}
\usepackage{graphics}
\usepackage{cite}
\usepackage{url}
\usepackage{hyperref}
\usepackage{dsfont}
\usepackage{epsfig}
\usepackage{euscript}
\usepackage{psfrag}
\usepackage{latexsym}
\usepackage{bbm}
\usepackage{color}
\usepackage{amstext}
\usepackage{wasysym}
\usepackage{cuted}
\usepackage{comment}
\usepackage{algorithm}
\usepackage{algpseudocode}
\usepackage[normalem]{ulem}
\usepackage{hyperref}
\usepackage{breqn}

\newcommand{\mR}{{\mathbb R}}

\newcommand{\bA}{{\mathbf A}}

\newcommand{\bH}{{\mathbf H}}

\newcommand{\bK}{{\mathbf K}}

\newcommand{\bM}{{\mathbf M}}

\newcommand{\bQ}{{\mathbf Q}}

\newcommand{\bS}{{\mathbf S}}

\newcommand{\bX}{{\mathbf X}}

\newcommand{\bZ}{{\mathbf Z}}

\newcommand{\bc}{{\mathbf c}}

\newcommand{\be}{{\mathbf e}}
\newcommand{\bff}{{\mathbf f}}

\newcommand{\bk}{{\mathbf k}}

\newcommand{\bq}{{\mathbf q}}
\newcommand{\bs}{{\mathbf s}}
\newcommand{\bu}{{\mathbf u}}
\newcommand{\bv}{{\mathbf v}}

\newcommand{\bx}{{\mathbf x}}
\newcommand{\by}{{\mathbf y}}
\newcommand{\bz}{{\mathbf z}}

\newtheorem{theorem}{Theorem} 
\newtheorem{definition}{Definition}
\newtheorem{lemma}{Lemma} 
\newtheorem{assumption}{Assumption}
\newtheorem{remark}{Remark}
\newtheorem{problem}{Problem}

\begin{document}

\begin{frontmatter}

\title{Safe Navigation in Dynamic Environments using Density Functions\thanksref{footnoteinfo}}

\thanks[footnoteinfo]{This paper was supported in part by the National Science Foundation under Grant No. NSF 2031573.}

\author[Aff1]{Sriram S. K. S. Narayanan}\ead{sriramk@clemson.edu},
\author[Aff2]{Joseph Moyalan}\ead{jmoyalan@ucmerced.edu},
\author[Aff1]{Umesh Vaidya\corauthref{cor1}}\ead{uvaidya@clemson.edu}
\corauth[cor1]{Corresponding author.}

\address[Aff1]{Department of Mechanical Engineering, Clemson University, Clemson, SC 29634, USA}
\address[Aff2]{Department of Mechanical Engineering, University of California, Merced, CA 95343, USA}


\begin{abstract}
This work presents a density-based framework for safe navigation in dynamic environments characterized by time-varying obstacle sets and time-varying target regions. We propose an analytical construction of time-varying density functions that enables the synthesis of a feedback controller defined as the positive gradient of the resulting density field. The primary contribution of this paper is a rigorous convergence proof demonstrating almost-everywhere safe navigation under the proposed framework, specifically for systems governed by single-integrator dynamics. To the best of our knowledge, these are the first analytical guarantees of their kind for navigation in dynamic environments using density functions. We illustrate the applicability of the framework to systems with more complex dynamics, including multi-agent systems and robotic manipulators, using standard control design techniques such as backstepping and inverse dynamics. These results provide a foundation for extending density-based navigation methods to a broad class of robotic systems operating in time-varying environments.
\end{abstract}

\end{frontmatter}

\section{Introduction}\label{sec:intro}
Safe navigation in dynamic environments is a fundamental challenge in robotics and autonomous systems \cite{hewawasam2022past, savkin2015safe}. The objective is to find a safe trajectory that the system must follow to reach a target (or track a target trajectory) while avoiding dynamic obstacles. Over the years, several methodologies have been developed to address these challenges effectively, such as sample-based methods, gradient-based methods, optimization-based methods, and reachable set computations \cite{hoy2015algorithms}.

Sample-based algorithms such as Rapidly-exploring Random Trees (RRT) and Probabilistic Roadmaps (PRM) are widely adopted in navigation tasks due to their flexibility and efficiency in high-dimensional spaces. RRT* improves upon RRT by ensuring asymptotic optimality, while variants like RRT-Connect enhance computational speed and scalability. Their application to dynamic environments has been explored in works such as \cite{lavalle1998rapidly, hsu2002randomized, likhachev2005anytime}, though these methods inherently lack formal safety guarantees. In contrast, gradient-based methods offer computational efficiency and are well-suited for real-time implementation, often employing artificial potential fields (APF), where attractive and repulsive forces guide navigation. The seminal work in \cite{khatib1986real} introduced APF for real-time obstacle avoidance, but such methods are susceptible to local minima and oscillatory behavior, especially in dynamic settings. To mitigate these issues, enhancements like the Virtual Force Field (VFF) \cite{borenstein1991vector} and the Navigation Function (NF) framework \cite{rimon1992exact} have been proposed. While navigation functions provide formal safety and convergence guarantees under specific topological constraints, their applicability is often limited by the requirement of constructing smooth diffeomorphic mappings to a sphere world, which may not be feasible in cluttered or highly dynamic environments. As an alternative, the Social Force Model (SFM) was originally introduced to simulate pedestrian motion \cite{helbing1995social}, treating agents as particles subject to social forces including attraction to goals and repulsion from obstacles and other agents. This model has been adapted to various robotic navigation tasks, particularly in scenarios involving human-robot interaction \cite{helbing2000simulating}. However, SFM-based approaches lack formal guarantees of safety or convergence, and their performance is sensitive to parameter tuning and the modeling accuracy of social interactions, limiting their robustness in complex or highly dynamic environments.


In recent years, Control Barrier Function (CBF) methods have emerged as a powerful approach for ensuring safety in dynamic environments \cite{ames2019control, igarashi2019time, hamdipoor2023safe}. For instance, in \cite{desai2022clf}, CBFs were combined with Control Lyapunov Functions (CLFs) in a quadratic programming framework to enable nonholonomic mobile robots to navigate safely through dynamic obstacles. Similarly, CBF-based controllers have been successfully applied to multi-agent systems for collision avoidance in complex, interactive settings \cite{jankovic2023multiagent, srinivasan2018control, lindemann2019control}. However, while CBFs provide formal safety guarantees, they do not ensure convergence to a goal, often necessitating augmentation with CLFs to achieve both safety and stability. Constructing suitable Lyapunov functions is itself a challenging task, and the construction of CBFs requires careful consideration of the system dynamics and safety specifications \cite{garg2024advances}. 

Reachability-based methods play a crucial role in ensuring safe navigation in dynamic environments by precomputing the set of all states a system can reach within a given time frame. Hamilton-Jacobi (HJ) reachability analysis is a verification method that computes the reach-avoid set, which encompasses the states from which a system can safely reach a target while adhering to time-varying constraints \cite{bansal2017hamilton}. This approach has been applied in various contexts. For instance, \cite{malone2017hybrid} demonstrated its utility in stochastic environments, combining HJ reachability with potential fields for dynamic obstacle avoidance. \cite{zhou2018efficient} and \cite{bajcsy2019efficient} refined these methods for efficient path planning and provably safe navigation in uncertain environments. Recent advancements, such as multi-time reachability by \cite{doshi2022hamilton}, further enhance its applicability in complex scenarios with time-varying obstacles and constraints. Despite its rigorous guarantees, a key limitation of HJ reachability methods is their high computational cost, which poses challenges in scaling to systems with high-dimensional state spaces.  

{\color{black}}
\begin{figure}[t]
    \centering
  \includegraphics[width=1\linewidth]{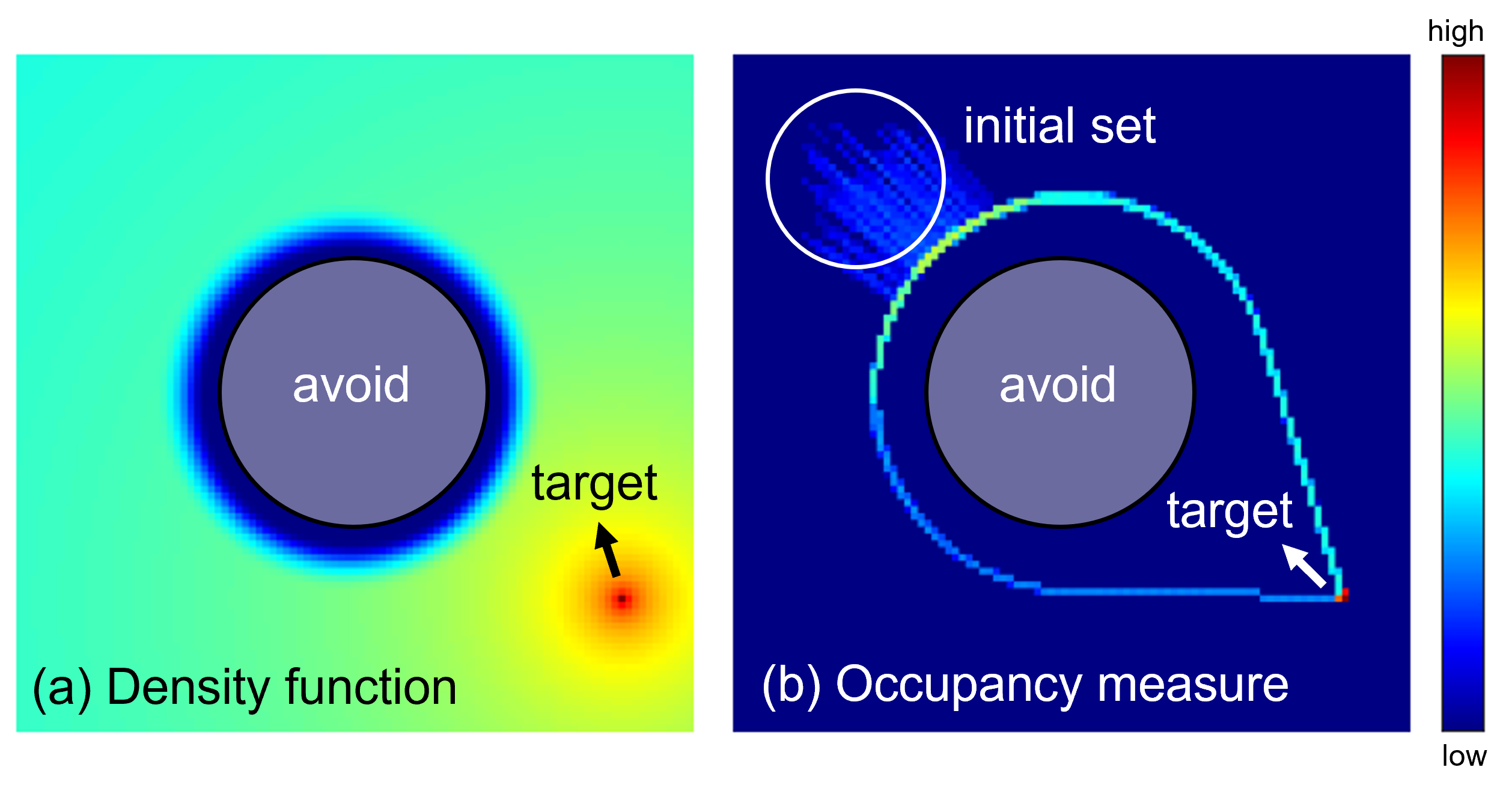}
  \caption{(a) Density function defined on an environment with a circular unsafe set and a point target, (b) Corresponding occupancy measure obtained using trajectories from 100 initial conditions sampled within the initial set. (color bar is in log scale)}
  \label{fig:dens_nav_diagram}
\end{figure}

Alternatively, the navigation problem can be formulated in the dual space of density, enabling the synthesis of feedback controllers that prioritize safety over optimality and scale efficiently to high-dimensional systems. Density functions have been shown to serve as both safety certificates and weaker notions of convergence, forming the basis for safe navigation control \cite{rantzer2004analysis, dimarogonas2007application, loizou2008density}. A convex formulation using the navigation measure was proposed in \cite{vaidya2018optimal}, and later extended to data-driven methods via linear transfer operators under safety constraints \cite{yu2022data, 10081458, moyalan2023off}. More recently, \cite{zheng2023safe} introduced an analytical density-based controller that jointly ensures obstacle avoidance and convergence. The key idea is to leverage the physically intuitive notion of occupation encoded by density functions, and therefore navigation, to design analytical feedback controllers that ensure almost-everywhere (a.e.) safe navigation. {\color{black} Note that time-varying obstacle or target sets are often addressed through optimization-based methods, which frame safety-critical control as a constraint satisfaction problem~\cite{ames2019control, bansal2017hamilton}. Similarly, density-based formulations have been extended to constrained nonlinear systems via optimization-based techniques; see, for instance, the control density function framework in~\cite{moyalan2024synthesizing}. In contrast, this work constructs an explicit feedback controller with rigorous guarantees for single integrator systems in time-varying environments. This remains a nontrivial challenge due to the need to encode dynamic safety and convergence guarantees in closed form. To the best of our knowledge, such a closed-form feedback construction for time-varying environments using a density-based approach has not been demonstrated before.
}

{\color{black} The main contributions of this paper are as follows. We provide a rigorous proof, including the explicit construction of a density-based feedback controller, for solving the navigation problem in dynamic environments with time-varying obstacle sets and a time-varying target set. This work extends the results presented in \cite{sriram2025_ACC}, which introduced a density-based framework for safe navigation in dynamic environments with applications to robotic systems. In particular, the theoretical guarantees for almost-everywhere safe navigation in the presence of both time-varying obstacles and time-varying target sets and their application to multi-agent systems and a two-link robotic arm are new to this paper. While the results are established for first-order integrator dynamics, to the best of the authors' knowledge, such a rigorous treatment does not currently exist in the literature. Importantly, proving these results for integrator dynamics enables us to extend the analytical construction procedure to a broader class of robotic systems using standard techniques such as inverse dynamics and backstepping. We demonstrate this capability in the application section by addressing navigation in dynamic environments for collision avoidance in a multi-agent system with agents' dynamics modeled as Dubin's car and safe trajectory tracking of a two-link robotic arm. Furthermore, this work opens new avenues for extending the results to a larger class of nonlinear systems with drift, using the framework of controlled density functions \cite{moyalan2024synthesizing}.}


The rest of the paper is organized as follows. In Section \ref{sec:prelims}, we introduce the preliminaries and illustrate the use of density functions for a.e. navigation in static environments. In Section \ref{sec:main_results}, we define the problem statements and provide the main results for a.e. safe navigation in dynamic environments. Specifically, we provide theorems to guarantee safety under dynamic obstacles (with static targets) and dynamic targets (with static obstacles). In Section \ref{sec:applications}, we show applications of the proposed approach with an integrator system in dynamic environments, multi-agent collision avoidance, and safe tracking for a two-link planar robotic arm. Finally, in Section \ref{sec:conclusions}, we provide concluding remarks for this work.
\section{Preliminaries and Notations}\label{sec:prelims}
\noindent {\bf Notations}: We use $\mathbb{R}^n$ to denote the $n$-dimensional Euclidean space. Let $\bx \in \mathbb{R}^n$ represent the system state vector and $\bu \in \mathbb{R}^m$ the control input vector. The workspace of the robot is denoted by a bounded set $\bX \subset \mathbb{R}^n$. Within this workspace, $\bX_0$, $\bX_T$, and $\bX_{u_k} \subset \bX$ for $k = 1, \ldots, L$ denote the initial set, target set, and the $k$-th unsafe region, respectively. A static target is represented by $\bx_T \in \bX_T$, while a time-varying target trajectory is denoted by $\bx_T(t)$. The overall unsafe set is defined as $\bX_u := \bigcup_{k=1}^L \bX_{u_k}$, and the safe set is $\bX_s := \bX \setminus \bX_u$. We use $\mathcal{C}^k(\bX)$ to denote the space of all $k$-times continuously differentiable functions on $\bX$, and define $\bX_1 := \bX \setminus \mathcal{B}_\delta$, where $\mathcal{B}_\delta$ is the $\delta$-neighborhood of the origin for arbitrarily small $\delta > 0$. The space of all measures on $\bX$ is denoted by $\mathcal{M}(\bX)$, and $m(\cdot)$ denotes the Lebesgue measure. The indicator function for a set $A \subset \bX$ is $\mathds{1}_A(\bx)$, and $\|\bx\|$ denotes the Euclidean norm of the vector $\bx$.

Density functions are a physically intuitive way to solve almost everywhere (a.e.) safe navigation (with respect to the Lebesgue measure) presented in Problem \ref{problem1} (defined in Section \ref{sec:main_results}). In this paper, we define safe trajectories for a system as the ones that have zero occupancy in the unsafe set $\bX_u$. The definition of occupancy used in this paper is as follows.
\begin{definition}[Occupancy of a set] Let $\bA\subset \bX$ be a measurable set. The occupancy of system trajectories $\bx(t)$ with initial condition $\bx$, in the set $\bA$ while traversing from the initial set $\bX_0\ni \bx$ to the target set $\bX_T$ is defined as 
\begin{align}
\mu(\bA):=\int_0^\infty \int_{\bX} {\mathds 1}_\bA(\bx(t))\mathds{1}_{\bX_0}(\bx)d\bx\;dt .\label{occupancy}
\end{align}
\end{definition}
The occupancy measure as defined in (\ref{occupancy}) was introduced in \cite{vaidya2018optimal} as a {\it navigation measure} for solving a.e. navigation problem for a discrete-time dynamical system. Using the physical interpretation of occupancy, it follows that if the measure is zero on a particular set, then that set will not be occupied and hence traversed by the system trajectories. Under the assumption that the measure is continuous with respect to Lebesgue and following the Radon–Nikodym theorem, one can define a density function. The construction of such a density function for safe navigation in a dynamic environment is the focus of this paper.

In Fig. \ref{fig:dens_nav_diagram}a, we show a plot of such a density function and the associated dynamics. We see that the system trajectories have zero occupancy on the unsafe set $\bX_u$ and a maximum occupancy in the target set $\bX_T$. So, by ensuring that the navigation density is zero on the unsafe set and maximum at the target set, it is possible to induce dynamics whereby the system trajectories will reach the desired target set while avoiding the unsafe set. It has been shown in our prior work~\cite{zheng2023safe} that the density-based framework does not suffer from the classical local minima issues associated with artificial potential field methods, thereby ensuring almost-everywhere convergence. We exploit this occupancy-based interpretation in the construction of analytical density functions for safe navigation in a dynamic environment consisting of time-varying obstacles and target sets.
\begin{definition}[Almost everywhere (a.e.) stability]  The target set $\bX_T$ of the system is said to be almost everywhere (a.e.) stable w.r.t. measure $\mu\in {\mathcal{M}}(\bX)$ if $\mu\{\bx\in \bX: \lim_{t\to \infty} \bx(t)\notin \bX_T\}=0.$
\label{def_aeuniformstable}
\end{definition}

Next, for safe navigation in dynamic environments, we first use the a.e. convergence criteria for time-varying systems introduced in \cite{masubuchi2021lyapunov}.
\begin{lemma}[\cite{masubuchi2021lyapunov}] \label{lemma_1} Given the time-varying system dynamics $\dot{\bx}(t) = \bff(t,\bx)$, if there exists a non-negative integrable function $\rho(t,\bx) \in \mathcal{C}^1(\mathbb{R}\times \bX_1,\mathbb{R})$ such that
\begin{subequations}
\begin{align}
    &\frac{\partial \rho}{\partial t} + \nabla\cdot \left(\bff(t,\bx)\rho(t,\bx)\right)> 0,\;a.e. \;(t,\bx)\in \mathbb{R} \times \bX_1, \label{lemma1_stat1}\\
    &\int_{\mathbb{R} \times \bX_1} \frac{1+\| \bff(t,\bx)\|}{1+\|\bx\|}\rho(t,\bx)\; d\bx \; dt < \infty, \label{lemma1_stat2}
\end{align}
\end{subequations}
then the system trajectories will converge to $\bX_T$ from almost all initial conditions (w.r.t. Lebesgue measure) inside $\bX_1$.
\end{lemma}

The results of Lemma \ref{lemma_1} provide the condition of a.e. convergence of time-varying systems and can be viewed as an extension of \cite{rantzer2001dual} to the time-varying setting. 

\begin{lemma}[\cite{masubuchi2021lyapunov}]\label{lemma_2} Consider the time-varying system dynamics $\dot{\bx}(t) = \bff(t,\bx)$, and a non-negative integrable function $\rho(t,\bx) \in \mathcal{C}^1(\mathbb{R}\times \bX_1,\mathbb{R})$. Let $s_{t}(t_0,\bx)$ be the solution of the system $\dot{\bx} = \bff(t,\bx)$. If $Z \subset \bX_1$ be a Borelian set, then for all $t_0 \le \tau \le t$, we can show that
    \begin{align}
        &\int_{s_t(t_0,\bZ)}\rho(t,\bx)d\bx - \int_{\bZ}\rho(t_0,\bx)d\bx = \nonumber \\
        &\int_{t_0}^t \int_{s_{\tau}(t_0,\bZ)}\left[\frac{\partial \rho(\tau,\bx)}{\partial \tau}+\left[\nabla\cdot(\bff \rho) \right](\tau,\bx) \right]\,d\bx\,d\tau. \label{eq:lemma2}
    \end{align}    
\end{lemma}
The results of Lemma \ref{lemma_2} can be viewed as an extension of classical Liouville results for an autonomous system with a time-invariant density function \cite{arnold1974equations}.

\subsection{ Safe Navigation in Static Environments}
In \cite{zheng2023safe}, an analytical construction of density function, $\rho(\bx)$, was provided for safe navigation in an environment consisting of static obstacle sets. Specifically, the density function proposed in \cite{zheng2023safe} can be used for obstacle avoidance while also satisfying the convergence to the target set properties. For a 2D single integrator system defined by $\dot{\bx}=\bu$, the control law given by the positive gradient of the density function, i.e., $\bu = \nabla \rho(\bx)$ will guarantee almost everywhere (a.e.) safe navigation, i.e., the system trajectories will converge to the target set $\bX_T$ while avoiding the unsafe set $\bX_u$. We provide an example to demonstrate the convergence and avoidance properties of such a controller.

\begin{exmp}[Static Safe Navigation] \label{ex:1}
Consider the integrator dynamics $\dot{\bx}=\bu$ with control law $\bu=\nabla \rho(\bx)$ (i.e., positive gradient of density function). The environment is defined with the target set at $\bx_T = [10,\; 0]$ with three circular unsafe sets $\bX_{u_1},\; \bX_{u_2}$ and $\bX_{u_3}$ each with radius $r=1$. The radius of the sensing region for each obstacle is given by $s_k$ as shown in Fig. \ref{fig:static_single_integrator}a. Note that we show two solution trajectories obtained with $s_{k_1}=2$ and $s_{k_2}=2.5$, respectively. It can be seen that both trajectories to the target while avoiding unsafe sets. 

Fig. \ref{fig:static_single_integrator}b shows the value of the density function $\rho(\bx)$ along the solution trajectory. Note that $\rho(\bx)>\theta>0 \; \forall \;t$, i.e., the trajectories never enter the unsafe set as $\rho(\bx)=0$ on the unsafe set. Further, $\rho(\bx)$ is the maximum near the target. This matches with the occupancy-based interpretation of density functions provided in \cite{zheng2023safe}. The main contribution of this work is to provide a solution to the a.e. safe navigation problem in a dynamic environment consisting of time-varying obstacles and target sets. 

\begin{figure}
    \centering
    \includegraphics[width = \linewidth]{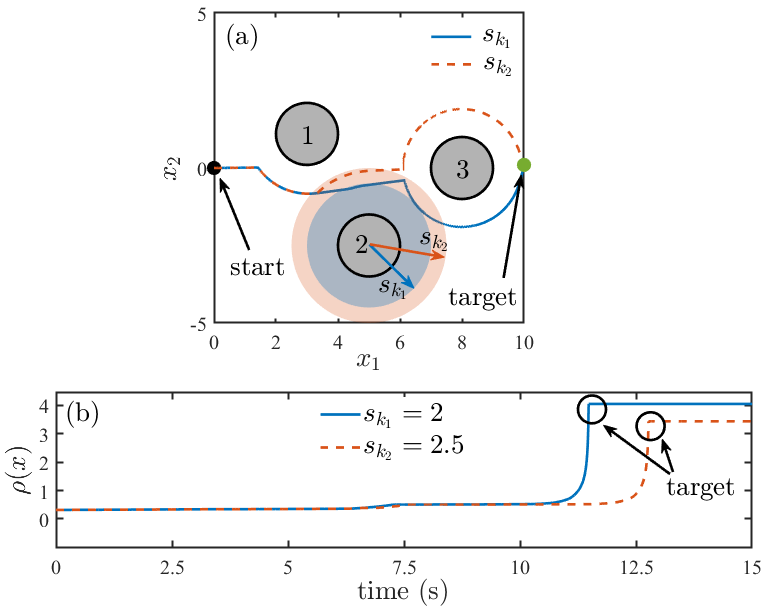}
    \caption{\textbf{Static Safe Navigation:} (a) Solution trajectory (red) obtained using the density-based controller and (b) the corresponding value of $\rho(\bx)$.}
    \label{fig:static_single_integrator}
\end{figure}
\end{exmp}

\section{Dynamic Safe Navigation} \label{sec:main_results}
In this section, we define the problem statement a.e. safe navigation with time-varying unsafe and target sets, provide the construction of time-varying density functions and develop the main results of this work in Theorem \ref{thm:time_varying_obs} and Theorem \ref{thm:time_varying_target}.
\begin{problem}[Dynamic Safe Navigation]\label{problem1} 
Consider the first-order integrator dynamics of the form
\begin{align}\label{sys1}
 \dot \bx=\bu.
 \end{align}
The objective is to design a feedback control input $\bu=\bk(t,\bx)$, possibly time-varying, to drive the trajectories of \eqref{sys1}
from a.e.(w.r.t. Lebesgue measure)\footnote{In the rest of the paper it is implicitly assumed that a.e. is w.r.t. Lebesgue measure unless stated otherwise.} initial condition from the initial set $\bX_0$ to a time-varying target set $\bx_T$ while avoiding time-varying unsafe set $\bX_{u_k}$ for $k=1,\ldots,L$. We assume that either the unsafe sets $\bX_{u_k}(t)$ or $\bx_T(t)$ are time-varying.  
\end{problem} 
\begin{remark} While the results are developed for first-order integrator dynamics, in our simulation section, we demonstrate how the results can be generalized to important classes of dynamics that arise in robotics applications such as Dubin's car model and for systems for which controller can be designed using inverse kinematics. 
\end{remark}

\begin{assumption} In the construction of the density function for a.e. navigation in a dynamic environment, it is assumed that the controller has information about the time-varying unsafe and target set. 
\end{assumption}
\begin{remark} While the theoretical results are developed for the case where the complete information of the dynamic environment is available in the construction of density functions, we will demonstrate through the simulation example how the theory can be applied for the cases where only local information is available.   
\end{remark}

\subsection{Construction of Time-Varying Density Functions}\label{sec:construction_nav_dens}

{\color{black} Building on the occupancy-based interpretation of density functions introduced in \cite{zheng2023safe}, we develop analytical expressions for time-varying density functions.} For each moving obstacle indexed by $k$, for $k = 1,\hdots L$ obstacles, we begin by defining its associated unsafe set $\bX_{u_k}(t)$, whose boundary is characterized by the zero-level set of a scalar function. Specifically, let $h_k(t,\bx): \mathbb{R}^n \to \mathbb{R}$ be a continuous scalar-valued function such that the boundary of $\bX_{u_k}(t)$ corresponds to the zero-level set $\{ \bx \in \bX : h_k(t,\bx) = 0 \}$. The unsafe set is then defined as
\begin{align}
\bX_{u_k}(t) := \{ \bx \in \bX : h_k(t,\bx) \leq 0 \}. \label{obstacle_region}
\end{align}
To ensure a well-posed obstacle representation, we assume that the level sets of $h_k(t,\bx)$ are connected. As an example, consider a circular obstacle whose center moves along the trajectory $\bc_k(t)$ with radius $r_k$; the corresponding unsafe region can be described as
\begin{align} 
\bX_{u_k}(t) = \{ \bx \in \bX : \|\bx - \bc_k(t)\| \leq r_k \}. \label{eqn:obstacles}
\end{align}
Surrounding each unsafe region, we define a sensing region $\bX_{s_k}(t)$, which delineates the area within which the robot is capable of detecting and responding to the obstacle. Let $s_k(t,\bx): \mathbb{R}^n \to \mathbb{R}$ be another continuous scalar function whose zero-level set defines the boundary of this sensing region. The sensing region is given by
\begin{align}
\bX_{s_k}(t) := \{ \bx \in \bX : s_k(t,\bx) \leq 0 \} \setminus \bX_{u_k}(t). \label{sensing_region}
\end{align}
Again, we assume that the level sets of $s_k(t,\bx)$ are connected. For circular obstacles, a corresponding circular sensing region may be described as
\begin{align} 
\bX_{s_k}(t) = \{ \bx \in \bX : \|\bx - \bc_k(t)\| \leq s_k \} \setminus \bX_{u_k}(t), \label{eqn:sensing}
\end{align}
where the sensing radius $s_k$ is strictly greater than the obstacle radius $r_k$. This sensing region is instrumental in extending the proposed navigation framework to settings where only local obstacle information is available. While the density formulation permits representation of arbitrarily shaped obstacles \cite{zheng2023safe}, we restrict our focus here to circular geometries for analytical tractability.

To encode the obstacle information, we define a smooth inverse bump function $\Psi_k(t, \bx)$, which captures the spatial profile of the unsafe set. This function is built from a family of smooth $\mathcal{C}^\infty$ functions. We first define an elementary smooth function $f$ as follows:
\begin{align} \label{eq:elementary_f}
f(\tau) = \begin{cases}
\exp{\left(-\frac{1}{\tau}\right)}, & \tau > 0 \nonumber \\
0, & \tau \leq 0
\end{cases},
\end{align}
for $\tau \in \mathbb{R}$. Using this, a smooth approximation to a step function is defined as
\begin{equation} 
\bar{f}(\tau) = \frac{(1 - \theta) f(\tau)}{f(\tau) + f(1 - \tau)} + \theta, \nonumber
\end{equation}
where $0 < \theta < 1$ is a positive scalar. To incorporate the geometry of both the obstacle and sensing region, we perform a change of variables using
\begin{align}
\phi_k(\bx - \bc_k(t)) = \bar{f}\left( \frac{\|\bx - \bc_k(t)\|^2 - r_k^2}{s_k^2 - r_k^2} \right). \label{ffunction}
\end{align}
This leads to the construction of the full inverse bump function
\begin{align}
\Psi_k(\bx - \bc_k(t)) := 
\begin{cases}
\theta, & \bx \in \bX_{u_k}(t), \\
\phi_k(\bx - \bc_k(t)), & \bx \in \bX_{s_k}(t), \\
1, & \text{otherwise}
\end{cases}. \label{inverse_bump}
\end{align}
\begin{remark}\label{remark_3}
    For notation convenience, we will simply write $\Psi_k(\bx-\bc_k(t))$ as $\Psi_k(t,\bx)$, however later in the proof of one of the main results, we will use this specific form of the inverse bump function $\Psi_k(t,\bx)$. When $\bx$ is outside the sensing region, i.e., $\bx \notin \bX_{s_k}$, we have $\Psi = 1$ and hence both the first and second derivatives of $\Psi$ w.r.t. $\bx$ are zero. Further, $\Psi_k(\bx,t)$ makes a smooth transition from $0<\theta<1$ to 1 inside the sensing region. $\alpha$, $\theta$, and $\bs_k$ are scalar tuning parameters that can be used to obtain trajectories with the desired behavior.
\end{remark}
The above construction of the function $\Psi_k$ is used to ensure that the system trajectory will have zero occupancy on the unsafe set. To ensure that the system dynamics are attracted to the target set, we introduce a distance function, $V(\bx)$. The distance function can be chosen to adapt to the geometry of the underlying configuration space of the system. For a Euclidean space with $\bx \in \mathbb{R}^n$, we pick $V(\bx) =\|\bx\|^2$ (if the target is assumed to be at the origin). Note that in Euclidean space, this distance function is sufficient, since the system dynamics are assumed to be integrators. The construction of an appropriate distance function for a general nonlinear system remains a challenge, but is not the focus of this paper.  
\begin{definition} [Dynamic unsafe set]  The navigation density function for the dynamic unsafe set and static target set is defined as 
\begin{align}
\rho_o(t,\bx)=\frac{\prod_{k=1}^L \Psi_k(t,\bx)}{(V(\bx)+\kappa)^\alpha}=\frac{\Psi(t,\bx)}{V_1(\bx)^\alpha}\label{density_fun1},
\end{align}
\end{definition}
where $V_1(\bx)=V(\bx)+\kappa$ for some constant $\kappa>0$ to ensure that the denominator is bounded away from zero.
For representing dynamic target sets, we next introduce a distance function, $V(t,\bx)$, which measures the distance from $\bx$ to the target set. The explicit time dependence of the distance function $V$ reflects the fact that the target set could be dynamic.  For a Euclidean space with $\bx \in \mathbb{R}^n$, we pick $V(t,\bx) =\|\bx-\bx_T(t)\|^2$, where $\bx_T(t)$ is the known dynamics of the target set. 
{\begin{definition} [Dynamic target set] The navigation density function for the case  where the target set is dynamic and the unsafe set is static is defined as 
\begin{align}
\rho_T(t,\bx)=\frac{\prod_{k=1}^L \Psi_k(\bx)}{(V(t,\bx)+\kappa)^\alpha}=\frac{\Psi(\bx)}{V_1(t,\bx)^\alpha}\label{density_fun2},
\end{align}
where $V_1(t,\bx)=V(t,\bx)+\kappa$ for some constant $\kappa>0$. Here, $\Psi_k(\bx)$ is as given in \eqref{inverse_bump} with $\bc_k(t)$ replaced with $\bc_k$ a constant independent of time. 
\end{definition}
The construction of this density function consists of two parts. The $\Psi(t,\bx)$ or $\Psi(\bx)$ captures the information of the unsafe set, and $V(\bx)$ encodes the target set.  The function $\Psi(t,\bx)$ is essentially a zero-one function, taking a small value $\theta$ inside the unsafe set and making a smooth transition from $\theta$ to one within the sensing region. This means that the information about the unsafe set is not known globally but only inside the sensing region (\ref{sensing_region}). This feature becomes especially important in the application of our framework to multi-agent collision avoidance, where each agent relies only on local information about neighboring agents (see Section \ref{sec:simulation_multiagent}).
\begin{remark}\label{remark2} Static unsafe set and static target set will be the special case of the construction procedure outlined above. In particular, for the case when the unsafe set is static, $\bc_k(t)$ can be replaced with $\bc_k$ (i.e., independent of time), signifying the center of the unsafe set.  Similarly, the static target set will be the special case of the time-varying target case.   
\end{remark}
\begin{figure}
    \centering
  \includegraphics[width=1\linewidth]{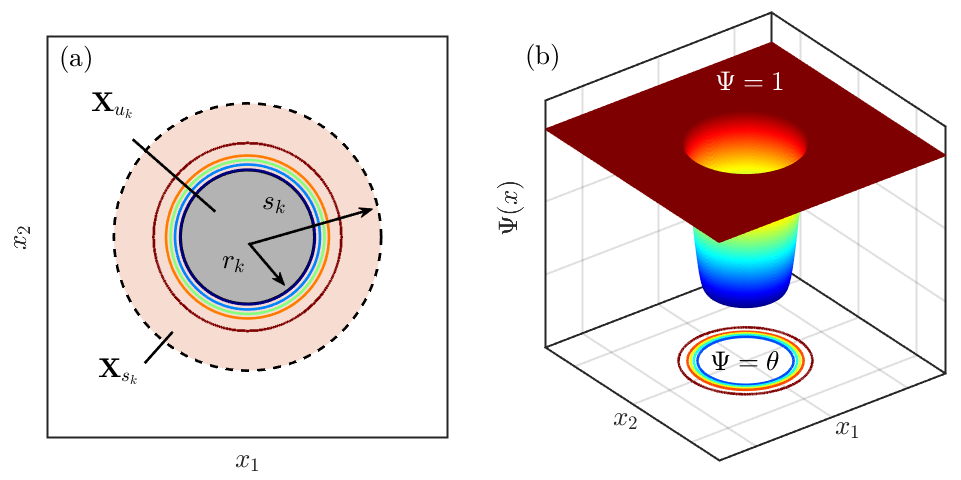}
  \caption{Inverse bump function $\Psi(\bx)$ (a) top view showing contours and (b) 3D view.}
  \label{fig:density_function}
\end{figure}

Next, we make the following technical assumptions, which will be used throughout the development of Lemma~\ref{lemma:ineq}, Theorems \ref{thm:time_varying_obs}, and \ref{thm:time_varying_target}. 

\begin{assumption}\label{assume_main} 
 \begin{enumerate}
\item[] 
\item [1.] We assume that the distance between the initial set, the unsafe sets, and the target sets are all bounded away from zero by some positive constant.
\item [2.] We assume that the obstacle sets $\bX_{u_k}$ for $k=1,\ldots,L$ are inside the bounded subset of $\bX_1$.
\item [3.] For the time-varying obstacle case, we assume that for all $k=1,\dots,L$, $ c_k(t)$ and $ \dot{c}_k(t)$ are continuous and bounded functions for all $t\ge0$.
\item [4.] In the sensing region, i.e., $\bx \in \bX_{s_k}$, we assume the following uniform bounds independent of $\bx$ and for all $t\ge 0$ and $j=1,\ldots,n$.
\begin{align*}
    &\left|\frac{\partial \Psi}{\partial t}\right|\le c_{\Psi_t}, \; \; \left| \frac{\partial \Psi}{\partial x_j}\right|\leq \bar c_{\Psi_x}, \; \left| \frac{\partial^2 \Psi}{\partial x_j^2}\right|\leq \bar c_{\Psi_{x^2}}.
\end{align*}
\item [5.] For all $\bx \in \bX_1$, $t\ge0$, and $j=1,\ldots,n$, we assume 
\begin{align*}&\frac{\partial^2 V}{\partial x_j^2}\leq \bar d_{V_{x^2}},\;\;\underline{d}_V \|\bx\|^2 \le V\leq \bar{d}_V \|\bx\|^2,\nonumber\\ &\underline{d}_{V_x}\|\bx\|\leq \left|\frac{\partial V}{\partial x_j} \right|\leq \bar d_{V_x}\|\bx\|,\\
&\underline{c}_x\le\|\bx\|\le \bar{c}_x, \quad \textrm{where} \;\;\; \underline{c}_x, \; \bar{c}_x>0.
\end{align*}
\item [6.] {\color{black} For the time-varying target case, we assume that $ \bx_T(t)$ and $\dot{\bx}_T(t)$ are continuous and bounded functions for all $t\ge0$.} 
\end{enumerate}
\end{assumption}
{\color{black}
\begin{remark}
The uniform bounds on the function $\Psi$ in the sensing region can be assumed as all the obstacle sets, and hence the sensing region is bounded. Furthermore, the function $\Psi$ makes a smooth transition from $\theta$ to one. Similarly, the bounds on the distance function $V$ assume that the distance function is bounded from above and below by a quadratic function of $\bx$.  
\end{remark}
}
The following lemma establishes key inequalities fundamental to the main results developed in this paper.
{\color{black}
\begin{lemma} \label{lemma:ineq}
    Given the system in \eqref{system_navigation} with the density function defined in \eqref{density_fun1}, and Assumption \ref{assume_main}, there exists a choice of parameters $\alpha$, $\beta$, and $\theta$, as a function of the scalar constants defined in Assumption \ref{assume_main}, such that the following inequalities are satisfied.
    \begin{subequations}
    \begin{align}
        &\nabla\cdot(\bk\rho_o) 
        \ge \frac{\alpha\beta n}{V^\alpha V_1^\alpha} \bigg( 
        (2\alpha+1) \left( \bar{d}_V (\bar{c}_x)^2 + \kappa \right)^{-2} 
       (\underline{d}_{V_x}\,\bar{c}_x\,\theta)^2 \nonumber \\
        &\quad \quad \quad - 4 (\underline{d}_V)^{-1} \bar{d}_{V_x} \bar{c}_{\Psi_x} 
        (\underline{c}_x)^{-1} 
        - \kappa^{-1} \bar{d}_{V_{x^2}} 
        - \alpha^{-1} \bar{c}_{\Psi_{x^2}} 
        \bigg),
        \label{eq:subs4} \\
        & \quad \; \left| \frac{\partial \rho_o}{\partial t} \right| 
        = \frac{1}{V_1^\alpha} 
        \left| \frac{\partial \Psi}{\partial t} \right| 
        \le \frac{1}{V_1^\alpha} c_{\psi_t}.
        \label{ineq:55}
    \end{align}\label{eq:x_sensing}
    \end{subequations}
    for all $\bx$ inside the sensing region, i.e, when $\bx \in \bX_{s_k}$. 
    \begin{subequations}
    \begin{align}
        &\nabla\cdot(\bk\rho_o) 
        \ge \frac{\alpha\beta n}{V^\alpha V_1^\alpha} \bigg( 
        (2\alpha+1)(\bar{d}_V \|\bx\|^2 + \kappa)^{-2} 
        \theta^2 (\underline{d}_{V_x})^2 \|\bx\|^2 \nonumber \\
        &\quad \quad \quad \quad \quad - (\underline{d}_V)^{-1} \bar{d}_{V_{x^2}} \|\bx\|^{-2} 
        \bigg),
        \label{eq:subs3_pre} \\
        & \quad \quad \frac{\partial \rho_o}{\partial t} = 0. \label{ineq:555}
    \end{align}\label{eq:x_not_sensing}
    \end{subequations}
    for all $\bx$ outside the sensing region, i.e., when $\bx \notin \bX_{s_k}$:
    \end{lemma}
    \begin{pf}
        The proof of this Lemma is deferred to the Appendix.
    \end{pf}
}
\subsection{Dynamic Almost Everywhere Safe Navigation using Time-varying Density Functions}
\noindent
This section presents the main theoretical results of the paper on dynamic safe navigation. We begin by introducing the construction of time-varying density functions for encoding dynamic unsafe and target sets. We propose explicit feedback controllers and prove their convergence and safety properties under mild assumptions. The section is organized into two parts: Section \ref{sec:dynamics_unsafe_set} focuses on avoiding a time-varying unsafe set with a static target, and Section \ref{sec:dynamics_target_set} addresses tracking a time-varying target while avoiding static obstacles.
 
\subsubsection{Dynamic Unsafe Set} \label{sec:dynamics_unsafe_set}
Given the construction of $\rho_o(t,\bx)$ in \eqref{density_fun1} for dynamic unsafe set, we design a controller $\bu = \bk(t,\bx)$ using the positive gradient of $\rho_o(t, \bx)$, i.e.,
\begin{align}
&\dot \bx =\bk(t, \bx) = \beta \nabla \rho_o(t, \bx) \nonumber \\ 
&=\beta \Biggl(-\frac{\alpha}{V_1^{\alpha+1}}\frac{\partial V}{\partial \bx}\prod_{k=1}^L \Psi_k(t, \bx)
+\frac{1}{V_1^\alpha}\frac{\partial }{\partial \bx}\prod_{k=1}^L \Psi_k(t, \bx)\Biggr)^\top.
\label{system_navigation}
\end{align}
where $\beta$ is a positive constant and $\rho_o$ is as given in (\ref{density_fun1}). The role of the positive constants $\beta$ and $\alpha$ will be clarified in the proof of the main theorems of this paper. Note that the dynamics of the obstacle is captured in $\bc_k(t)$ (trajectory of obstacle $k$) which is used in the construction of $\Psi_k(t,\bx)$.
\begin{remark}The system dynamics for the dynamic unsafe set given by  (\ref{system_navigation}) is locally asymptotically stable. 
Following Assumption \ref{assume_main}.1, the system dynamics in the small neighborhood of the origin for a time-varying unsafe set is reduced to   
\begin{align*}
\dot \bx=\beta \Biggl(-\frac{\alpha}{V_1(\bx)^{\alpha+1}}\frac{\partial V}{\partial \bx}
\Biggr)^\top.
\end{align*}
Let $V(\bx)$ be the distance function, we have 
\begin{align*}
\dot V=\frac{\partial V}{\partial \bx}\beta \Biggl(-\frac{\alpha}{V_1^{\alpha+1}}\frac{\partial V}{\partial \bx}
\Biggr)^\top<0.
\end{align*}
Then $V(\bx)$ becomes the Lyapunov function. The local stability then follows using results from \cite{Khalil:1173048}. With no loss of generality, we can assume that the $B_\delta$ is a neighborhood of the origin where the system is locally stable. 
\end{remark} 

The following theorem establishes conditions for safe navigation in the presence of a dynamic unsafe set (with a static target set).


\begin{theorem}[Dynamic Unsafe Set]
\label{theorem_main}Under Assumption \ref{assume_main}, the system given in \eqref{system_navigation} with the density function defined in \eqref{density_fun1} will solve the a.e. navigation problem as stated in Problem \ref{problem1} with time-varying unsafe sets and static target sets if Lemma \ref{lemma:ineq} is satisfied.
\label{thm:time_varying_obs}
\end{theorem}
\begin{pf} The proof is structured in two parts. In the first part, concerning convergence, we invoke Lemma~\ref{lemma_1} to demonstrate that the dynamical system defined in \eqref{system_navigation} converges to the target set $\bX_T$ in an almost-everywhere sense. In the second part, addressing avoidance, we employ Lemma~\ref{lemma_2} to establish that the system avoids the unsafe set $\bX_u$ for a sufficiently small choice of the parameter $\theta$.

\textbf{Convergence:}
   We consider the closed-loop dynamics given by \eqref{system_navigation} and the density function for dynamic obstacles, $\rho_o$, defined in \eqref{density_fun1}. To show that the trajectories of this system converge to the target $\bX_T$ from almost all initial conditions, we will utilize Lemma~\ref{lemma_1} and prove that the following inequalities are satisfied.
\begin{subequations}
    \begin{align}
        &\nabla\cdot(\bk(t,\bx)\rho_o(t,\bx)) + \frac{\partial \rho_o}{\partial t} > 0,\;a.\,e. \; (t,\bx)\in \mathbb{R} \times \bX_1, \label{eq:sub1}\\
        &\int_{\mathbb{R} \times \bX_1} \frac{1+\| \bk(t,\bx)\|}{1+\|\bx\|}\rho_o(t,\bx)\; d\bx\; dt < \infty. \label{eq:sub2}
    \end{align}
\end{subequations}
 {\bf Validity of} \eqref{eq:sub1}: We will show that each term on the left-hand side of \eqref{eq:sub1} is greater than zero. We will first expand the divergence term, $\nabla\cdot(\bk(t,\bx)\rho_o(t,\bx))$ as 
    \begin{align}
        \nabla\cdot(\bk\rho_o)= \beta\left(\frac{\partial \rho_o}{\partial \bx}\frac{\partial \rho_o}{\partial \bx}^\top + \rho_o(t,\bx)\sum_{j=1}^n \frac{\partial^2\rho_o}{\partial x_j^2}\right).\label{divergence_term}
    \end{align}
    We know that $\rho_o(t,\bx)>0$ and each term in \eqref{divergence_term} can be expanded as follows:
    \begin{subequations}    
    \begin{align}
    &\frac{\partial \rho_o}{\partial \bx}\frac{\partial \rho_o}{\partial \bx}^\top 
    = \sum_{j=1}^n \left( \frac{\partial \rho_o}{\partial x_j}
    \frac{\partial \rho_o}{\partial x_j} \right) \nonumber\\
    &= \sum_{j=1}^n \bigg( 
    \frac{\alpha^2}{V^{\alpha+1}V_1^{\alpha+1}}\Psi^2 
    \left(\frac{\partial V}{\partial x_j}\right)^2 
    + \frac{1}{V^{\alpha}V_1^{\alpha}} 
    \left(\frac{\partial \Psi}{\partial x_j}\right)^2 \nonumber\\
    &\quad - \frac{\alpha}{V^{\alpha}V_1^{\alpha+1}}\Psi 
    \frac{\partial V}{\partial x_j} \frac{\partial \Psi}{\partial x_j}
    - \frac{\alpha}{V^{\alpha+1}V_1^{\alpha}}\Psi 
    \frac{\partial V}{\partial x_j} \frac{\partial \Psi}{\partial x_j} \bigg),
    \label{divergence_part1}\\
    &\rho_o\frac{\partial^2\rho_o}{\partial x_j^2} 
    = \frac{\alpha \Psi}{V^\alpha V_1^{\alpha}} \bigg(
    \frac{\alpha+1}{V_1^2} \Psi 
    \left(\frac{\partial V}{\partial x_j}\right)^2 
    - \frac{1}{V_1} \Psi 
    \frac{\partial^2 V}{\partial x_j^2} \nonumber\\
    &\quad \quad \quad - \frac{2}{V_1} \frac{\partial \Psi}{\partial x_j} 
    \frac{\partial V}{\partial x_j} 
    + \frac{1}{\alpha} \frac{\partial^2 \Psi}{\partial x_j^2} \bigg).
    \label{divergence_part2}
\end{align}
\end{subequations}
Therefore, using \eqref{divergence_part1} and \eqref{divergence_part2}, we can write \eqref{divergence_term} as
\begin{align}
    \nabla\cdot(\bk\rho_o) =\; &\frac{\alpha\beta}{V^\alpha V_1^\alpha} 
    \sum_{j=1}^n \Bigg( 
    \left( \frac{\alpha}{VV_1} + \frac{\alpha+1}{V_1^2} \right)\Psi^2 
    \left( \frac{\partial V}{\partial x_j} \right)^2 \nonumber\\
    &- \left( \frac{3}{V_1} + \frac{1}{V} \right)\Psi 
    \frac{\partial V}{\partial x_j} \frac{\partial \Psi}{\partial x_j} 
    - \frac{\Psi^2}{V_1} \frac{\partial^2 V}{\partial x_j^2} \nonumber\\
    &+ \frac{1}{\alpha} \left( 
    \Psi \frac{\partial^2 \Psi}{\partial x_j^2} + 
    \left( \frac{\partial \Psi}{\partial x_j} \right)^2 \right) 
    \Bigg).
    \label{divergence_final_term}
\end{align}
\underline{Inside the sensing region}, i.e, when $\bx \in \bX_{s_k}$, using Assumption ~\ref{assume_main}.2, ~\ref{assume_main}.4, and \ref{assume_main}.5, we arrive at the following inequalities (see \eqref{eq:x_sensing} in Lemma \ref{lemma:ineq}):
\begin{subequations}
\begin{align}
    &\nabla\cdot(\bk\rho_o) 
    \ge \frac{\alpha\beta n}{V^\alpha V_1^\alpha} \bigg( 
    (2\alpha+1) \left( \bar{d}_V (\bar{c}_x)^2 + \kappa \right)^{-2} 
   (\underline{d}_{V_x}\,\bar{c}_x\,\theta)^2 \nonumber \\
    &\quad \quad \quad - 4 (\underline{d}_V)^{-1} \bar{d}_{V_x} \bar{c}_{\Psi_x} 
    (\underline{c}_x)^{-1} 
    - \kappa^{-1} \bar{d}_{V_{x^2}} 
    - \alpha^{-1} \bar{c}_{\Psi_{x^2}} 
    \bigg), \nonumber \\
    & \quad \; \left| \frac{\partial \rho_o}{\partial t} \right| 
    = \frac{1}{V_1^\alpha} 
    \left| \frac{\partial \Psi}{\partial t} \right| 
    \le \frac{1}{V_1^\alpha} c_{\psi_t}. \nonumber
\end{align}
\end{subequations}

\underline{Outside the sensing region}, i.e., when $\bx \notin \bX_{s_k}$, using Assumption~\ref{assume_main}.4 and \ref{assume_main}.5, and the fact that $\frac{\partial \Psi}{\partial x_j}=0$, and $\frac{\partial^2 \Psi}{\partial x_j^2}=0$, we get the following conditions (see \eqref{eq:x_not_sensing} from Lemma \ref{lemma:ineq}):
\begin{subequations}
\begin{align}
    &\nabla\cdot(\bk\rho_o) 
    \ge \frac{\alpha\beta n}{V^\alpha V_1^\alpha} \bigg( 
    (2\alpha+1)(\bar{d}_V \|\bx\|^2 + \kappa)^{-2} 
    \theta^2 (\underline{d}_{V_x})^2 \|\bx\|^2 \nonumber \\
    &\quad \quad \quad \quad \quad - (\underline{d}_V)^{-1} \bar{d}_{V_{x^2}} \|\bx\|^{-2} 
    \bigg), \nonumber \\
    & \quad \quad \frac{\partial \rho_o}{\partial t} = 0. \nonumber
\end{align}
\end{subequations}
Now using the results of Lemma \ref{lemma:ineq}, the parameters $\alpha$, $\beta$, and $\theta$ can always be chosen to satisfy the conditions in \eqref{eq:x_sensing} and \eqref{eq:x_not_sensing}. 

{\bf Validity of} \eqref{eq:sub2}:
Based on Assumptions~\ref{assume_main}.4 and \ref{assume_main}.5, for all $\bx \in \bX_1$, we can have the following upper and lower bounds on each element of the vector field $\bk(t,\bx)$ denoted by $k_j(t,\bx)$ for $j=1,\dots,n$.
    \begin{subequations}\label{eq:subs56}
        \begin{align}
            &-k_1\le k_j(t,\bx)  \le k_1 \quad \forall \;\bx \; \in \bX_{s_k}, \label{eq:subs6}\\
            &-k_2 \le k_j(t,\bx)  \le k_2 \quad \forall \;\bx \; \notin \bX_{s_k}, \label{eq:subs5}
        \end{align}
    \end{subequations}
where $k_1$ and $k_2$ are scalar bounds defined as follows:
\begin{align*}
    &k_1 = \beta \left( \kappa^{-(\alpha+1)}\bar{d}_{V_x}\|\bx\|+\kappa^{-\alpha}\bar{c}_{\Psi_x}\right), \\
    &k_2 = \beta\alpha  \kappa^{-(\alpha+1)}\bar{d}_{V_x}\|\bx\|.
\end{align*}
 \underline{Inside the sensing region}, i.e., when $\bx \in \bX_{s_k}$, we know that $\|\bx\| \leq \bar{c}_x$. Therefore, by using this fact and the bounds in \eqref{eq:subs6}, we can infer that \eqref{eq:sub2} is satisfied when $\bx \in \bX_{s_k}$.

 \underline{Outside the sensing region}, i.e., when $\bx \notin \bX_{s_k}$, using \eqref{eq:subs5}, we get the bounds on $\|\bk(t,\bx)\|$ as follows:
 $$\|\bk(t,\bx)\| \; \le \sqrt{n}\,\alpha \beta \kappa^{-(\alpha+1)}\bar{d}_{V_x}\|\bx\|.$$
 Also, when $\bx \notin \bX_{s_k}$,
 $\rho(t,\bx) = V(\bx)^{-\alpha}$ and
 $V(\bx) \le \bar{d}_V\|\bx\|^2$. Therefore,
 \begin{align}
     \frac{1+\| \bk(t,\bx)\|}{1+\|\bx\|}\rho(t,\bx) \le \frac{\sqrt{n}\alpha \beta (\kappa)^{-(\alpha+1)}\bar{d}_{V_x}\|\bx\|}{(1+\|\bx\|)(\bar{d}_V\|\bx\|^2)^\alpha}. \label{eq:inetgral_ineq}
 \end{align}
 We observe that when $\bx \notin \bX_{s_k}$, as $\|\bx\| \rightarrow \infty$, the numerator in \eqref{eq:inetgral_ineq} goes to infinity linearly whereas the denominator goes to infinity at a rate of $(2\alpha+1)$. Therefore, we can conclude that $\eqref{eq:inetgral_ineq}<\infty$ for all $\bx \notin \bX_{s_k}$ and hence \eqref{eq:sub2} is satisfied. This concludes the convergence proof.
 
\textbf{Avoidance:}    
    In the second part of the proof, we show that the system trajectories with respect to $\bk(t,\bx)$ obtained from \eqref{system_navigation} avoid the unsafe set $\bX_u$. Utilizing Theorem~\ref{lemma_2}, we can write the evolution of the densities of the states along system trajectories as follows:
    \begin{align}
        &\int_{s_t(t_0,\bZ)}\rho_o(t,\bx)d\bx - \int_{\bZ}\rho_o(t_0,\bx)d\bx = \nonumber \\
        &\int_{t_0}^t \int_{s_{\tau}(t_0,\bZ)}\left[\frac{\partial \rho_o(\tau,\bx)}{\partial \tau}+\left[\nabla\cdot(\bk\rho_o) \right](\tau,\bx) \right]\;d\bx\; d\tau>0. \label{eq:proof_thm1}
    \end{align}
    This proof is done through the method of contradiction. First, we notice that the above quantity is greater than zero, following the previous convergence proof, where we showed that (\ref{eq:sub1}) is positive.  For a given $t_0 \ge 0$, let there exists an initial condition $\bx_0 \in \bX_0$ such that the solution of the vector field, $s_T(t_0,\bx_0) \in \bX_u$ for some $T>t_0$ and $s_t(t_0,\bx_0) \in \bX_1$ for $t \in [t_0,T]$. Let $\bZ \subset \bX_0$ be a positive Lebesgue measure set such that  $s_T(t_0,\bZ) \in \bX_u$ for some $T>t_0$ and $s_t(t_0,\bZ) \in \bX_1$ for $t \in [t_0,T]$. Since $\bZ \subset \bX_0$, utilizing Assumption~\ref{assume_main}.1 and from the construction of $\rho_o$, we have
    \begin{align}
        \int_{s_T(t_0,\bZ)}\rho_o(T,\bx)d\bx&=\theta\int_{s_T(t_0,\bZ)}\frac{1}{V_1(\bx)^{\alpha}}\;d\bx   \;, \label{eq:proof_23}\\ \int_{\bZ}\rho_o(t_0,\bx)d\bx &= \int_{\bZ}\frac{1}{V_1(\bx)^{\alpha}}\;d\bx .\label{eq:proof_24} 
    \end{align}
 As $\bZ$ is assumed to be positive Lebesgue measure set, we have (\ref{eq:proof_24}) greater than zero and since $\theta$ can be chosen to be a sufficiently small but positive constant, we have the difference, $\eqref{eq:proof_23}-\eqref{eq:proof_24}<0$, contradicting (\ref{eq:proof_thm1}). This concludes the proof for avoidance and thus completes the proof of Theorem~\ref{thm:time_varying_obs}.
\end{pf}
}
\subsubsection{Dynamic Target Set} \label{sec:dynamics_target_set}
Similarly, for tracking the time-varying target with static unsafe sets, we propose the following dynamics. 
\begin{align}
\dot \bx=\bk(\bx,t)=\beta \nabla\rho_T(t,\bx)+\dot{\bx}_T(t),\label{dynamics_target}
\end{align}
where again $\rho_T$ is as defined in \eqref{density_fun2} and $\beta$ is a positive constant. Notice that for the time-varying target case, we have system dynamics forced by $\dot \bx_T(t)$. 

\begin{remark}
The closed-loop system in \eqref{dynamics_target} will asymptotically track the target trajectory, $\bx_T(t)$, for all initial conditions $\bx$ in the small neighborhood of $\bx_T(t)$. For the time-varying target case, following Assumption \ref{assume_main}.1, the system dynamics in the small neighborhood of $\bx_T(t)$ is reduced to 
\begin{align}
    \dot{\bx} &= \beta \Biggl(-\frac{\alpha}{V_1(\bx)^{\alpha+1}}\frac{\partial V}{\partial \bx}
\Biggr)+\dot{\bx}_T \nonumber \\   
&= -\frac{2\alpha\beta}{(\|\bx-\bx_T\|^2+\kappa)^{\alpha+1}}(\bx-\bx_T) +\dot{\bx}_T \label{local_stability_target}
\end{align}
substituting $\be = \bx - \bx_T$ and $\dot{\be} = \dot{\bx} - \dot{\bx}_T$ in \eqref{local_stability_target}, we get
\begin{align}
    \dot{\be} = &-K_p\, \be, \label{error_states} \\
    K_p := &\frac{2\alpha\beta}{(\|e\|^2+\kappa)^{\alpha+1}} \cdot \nonumber
\end{align}
Since $K_p$ is uniformly bounded away from zero, we can conclude from \eqref{error_states} that $\be \rightarrow 0$ as $t \rightarrow \infty$. Therefore, the solution of the closed-loop system in \eqref{dynamics_target} will track the desired trajectory $\bx_T$ for all initial conditions starting inside the small neighborhood of $\bx_T$. 
\end{remark}
The following theorem establishes conditions for safe navigation in the presence of a dynamic target set (with static unsafe sets).
{\color{black}
\begin{theorem}[Dynamic Target Set]
\label{theorem_main2}Under Assumption \ref{assume_main} and Lemma \ref{lemma:ineq}, the dynamical system given in \eqref{dynamics_target} with the density function defined in \eqref{density_fun2} will solve the a.e. navigation problem as stated in Problem \ref{problem1} with static unsafe sets and time-varying target sets.
\label{thm:time_varying_target}
\end{theorem}
\begin{pf} Consider the change of coordinates given by $\by(t)=\bx-\bx_T(t)$ for the time-varying target case system in (\ref{dynamics_target}). Therefore, 
\begin{align}
    \dot \by=\beta \nabla \rho_T(t,\by+\bx_T(t)).\label{eq:time_vraying_target_dyn}
\end{align}
Now recalling the definition of $\rho_T(t,\bx)$ and using the fact that $V(\bx,t)=\|\bx-\bx_T(t)\|^2$, it follows that 
\begin{align}
    \rho_T(t,\by+\bx_T(t)) = \frac{ \Psi(\by+\bx_T(t))}{V(\by)^\alpha}\cdot \label{eq:time_vraying_target_proof_1}
\end{align}
Now, recalling Remark~\ref{remark_3}, we can rewrite \eqref{eq:time_vraying_target_proof_1} as follows:
\begin{align*}
    \rho_T(t,\by+\bx_T(t)) &= \frac{ \Psi(\by-(-\bx_T(t)))}{V(\by)^\alpha}= \rho_o(t,\by).
\end{align*}
 Therefore, we can rewrite \eqref{eq:time_vraying_target_dyn} as follows:
\begin{align}
    \dot \by=\beta \nabla \rho_o(t,\by).\label{eq:time_vraying_target_dyn_2}
\end{align}
Now, using Theorem~\ref{theorem_main} and Lemma \ref{lemma:ineq}, we can show that the solution of \eqref{eq:time_vraying_target_dyn_2} will converge to the target at the origin (in the $\by$-coordinates) while avoiding unsafe sets, since $\by \in \bX_1$ for all $t\ge0$ (Assumption~\ref{assume_main}.5). The appropriate range of $\alpha$ and $\beta$ are given by \eqref{alpha_choice} and \eqref{beta_choice} respectively. This ensures that the solution of \eqref{dynamics_target} will converge to $\bx_T(t)$ in the $\bx$-coordinates while avoiding unsafe sets.
\end{pf}
}
\begin{remark} \label{remark_tuning} The tuning parameters in the design of density functions are $\alpha$, $\beta$, $\theta$, and $s_k(\bx)$. The tuning of $\alpha$ depends on the rate of convergence of the trajectories, while $\beta$ is used to speed up the system dynamics, and $\theta$ determines the degree of safety (occupancy inside the unsafe set). In practice, $\alpha \in [0.1,\; 1]$, $\beta \in [1,\; 10]$ and $\theta \in [0.01,\; 0.1]$ works well in simulation. The tuning of $s_k(\bx)$ is physically intuitive, as it signifies the sensing region. Hence, a sensing region that encompasses the unsafe set with a sufficiently curved convex set has worked in the simulations.
\end{remark}

\section{Case Studies}
\label{sec:applications}
In this section, we present simulation results to verify the theoretical framework developed in this paper. All the simulations are conducted on an Intel Core i9-12900K CPU and 32 GB of RAM with a simulation timestep of \( 0.1 \) s. Implementation details can be found at \url{https://github.com/sriram-2502/time_varying_density}.
\subsection{Dynamic Unsafe Set}
The first example is for time-varying obstacle sets. 
Given the initial condition $\bX_0 = [0, \;0]$ and a static target $\bx_T = [10, \;0]$, the objective is to converge to the target while avoiding time-varying unsafe sets ($\bX_{u_k}$ for $k=1,\hdots,4$). The trajectory of the center $\bc_k(t)$ of each obstacle is defined as follows
\begin{align}
    &c_1(t) = \bigl[2, \; 0.25t \bigr], \quad c_2(t) = \bigl[4, \; 7-0.2t \bigr] \nonumber \\
    &c_3(t) = \bigl[6+0.1\sin(t), \; 0.15t-6 \bigr], \quad c_4(t) = \bigl[8, \; 5-0.12t \bigr]. \nonumber
\end{align}
Each obstacle is modeled as a circular disk with a radius $r_k = 0.75$ with the sensing radius of $s_k = 1.5$ as defined in \eqref{eqn:obstacles} and \eqref{eqn:sensing} respectively. We use Theorem \ref{theorem_main} to define the density-based controller for this system.
For this example we add input constraints, $\bu \in [-\bu_{\max}, \bu_{\max}]$ where $\bu_{\max}$ is the bound on control. Without formality, we constrain the control when $||\bu||_{\infty} > \bu_{\max}$ by normalizing the control as $\left(\frac{\bu}{||\bu||_\infty}\right)\bu_{\max}$.  This saturated version is applied to demonstrate feasibility. For a more principled treatment of control constraints under the density-based paradigm, we refer the reader to \cite{moyalan2024synthesizing}.

\begin{figure}
    \centering
    \includegraphics[width = \linewidth]{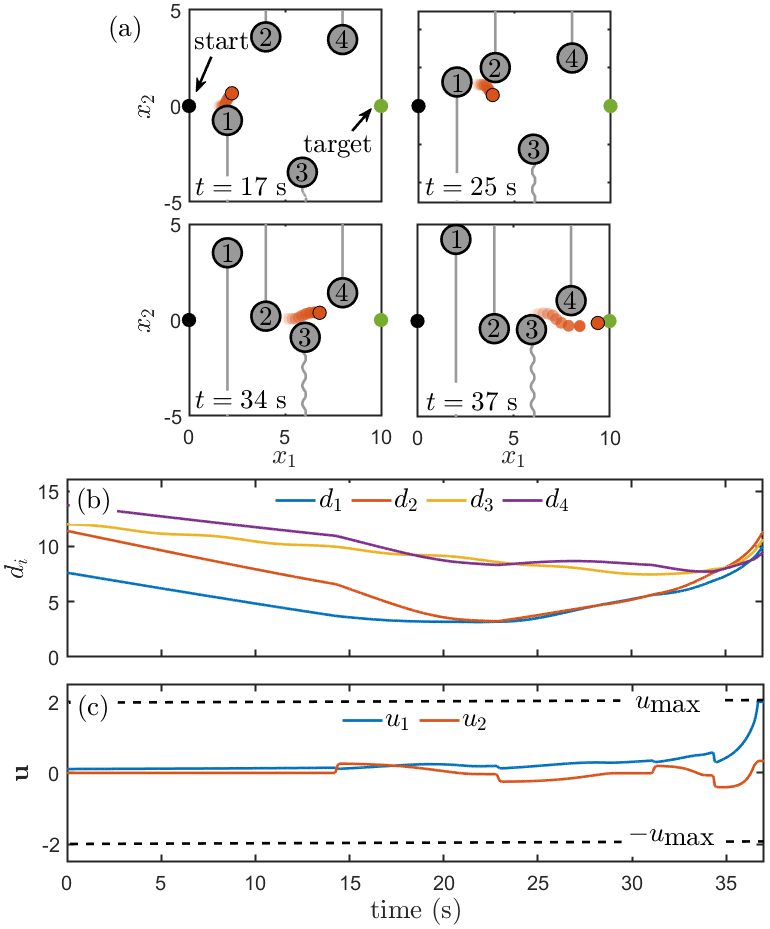}
    \caption{\textbf{Dynamic Unsafe Set:} Snapshots of the system trajectory (red) converging to the target (green) while avoiding time-varying obstacles (gray). }
    \label{fig:time_varying_obs}
\end{figure}

Fig. \ref{fig:time_varying_obs}a shows snapshots of the system trajectories avoiding each obstacle and converging to the target. The trajectories avoid each obstacle $\bX_{u_k}$ (labeled $1,\hdots,4$) at $t=17$ s, $20$ s, $34$ s, and $37$ s, respectively. 

Fig. \ref{fig:time_varying_obs}b shows the distance between the system trajectory and obstacles $d_k = \|\bx-\bX_{u_k}\|$ over time. It can be seen that this distance is always greater than zero for all time, indicating that the system trajectories never enter the unsafe set. The corresponding control inputs are shown in \ref{fig:time_varying_obs}c. Note that the control inputs stay within the bounds defined by $\bu_{\textrm{max}}=2$. 



\subsection{Collision Avoidance in Multi-agent Systems}\label{sec:simulation_multiagent}
In this section, we demonstrate the application of the proposed density-based controller to a multi-agent collision avoidance problem. Let $\bx_j(t) \in \mathbb{R}^n$ for $j = 1, \ldots, L$ denote the state of the $L$ agents. Define $\bX_{u_j}$ as the exclusion region associated with agent $\bx_j$, within which no other agent is permitted to enter. This region may represent either the physical space occupied by the agent, in which case $\bx_j$ is the geometric center of $\bX_{u_j}$, or a safety buffer (e.g., a collision-avoidance bubble) surrounding the agent. For the purposes of this work, we assume that $\bX_{u_j}$ is a circular region as defined in \eqref{eqn:obstacles}.

We further define the sensing region around each agent $j$ as $\bX_{s_j}(t)$, which is a neighborhood of $\bX_{u_j}(t)$ where other agents, $\bx_k(t)$ for $k = 1, \ldots, L$, $k \neq j$, can detect the presence of agent $j$. Within this sensing region, agents are capable of perceiving each other's presence and adjusting their control inputs accordingly. The sensing region is also modeled as a circular set and defined as in \eqref{eqn:sensing}.

In the following, we provide the construction of density functions for each agent. Let $\rho_{o_j}(t,\bx_j)$ be the density function of $j^{th}$ agent. We have the following construction of the density function:
\begin{align}
\rho_{o_j}(t,\bx)=\frac{\prod_k \Psi_k(t,\bx_j)}{\left(V_j(\bx)+\kappa_j\right)^{\alpha_j}} \cdot \label{eq:mutliagent_density}
\end{align}
for $k = 1, \ldots, L$, $k \neq j$, $\alpha_j>0$ and $\kappa_j>0$. Here $\Psi_k(t,\bx_j)$ is the inverse bump function for agent $j$ which encodes the other agent $k$ as an obstacle, and is defined as follows:
\begin{align}
{\color{black}\Psi_k(t,\bx_j)} := \begin{cases}
    {\color{black}\theta_k}, & \bx_j \in \bX_{u_k}(t)  \\
    \phi_k(\bx_j-\bx_k(t)), & \bx_j \in \bX_{s_k}(t)\\
    1, & \rm{otherwise}
\end{cases}.  \label{multiagentconst_2}
\end{align} 
The function $\phi_k(\bx_j-\bx_k(t))$ is as defined in (\ref{ffunction}). The function $V_j(\bx_j)$ encodes the target information $\bx_{T_j}$ for the agent $\bx_j$ and can be defined as follows:
\begin{equation}
    V_j(\bx_j) = \frac{1}{\|\bx_j-\bx_{T_j}\|^2} \cdot
\end{equation}
The regions, $\bX_{u_k}(t)$ and $\bX_{s_k}(t)$ play the same role as the regions defined in (\ref{obstacle_region}) and (\ref{sensing_region}) where the other agents $\bx_k(t)$ act as a time-varying obstacle for the agent $\bx_j$. The dynamics of the $j^{th}$ agent is then given by
\begin{align}
\dot \bx_j=\beta \nabla \rho_j(t,\bx_j).
\label{eq:multiagent_controller}
\end{align}

\begin{remark} It is important to emphasize that while the construction of the density function $\rho_j(t,\bx_j)$ in \eqref{eq:mutliagent_density} involves the knowledge of all the agents, the construction of function, $\Psi_k$ for each agent is such that the individual agents do not need to know the states of other agents until the sensing regions of the two agents collide.  Hence, the collision avoidance dynamics can be executed in a distributed manner without having access to the global information of all the agents. 
\end{remark}

Next, we set up an example to simulate an intersection where the agents have to avoid collision with each other while reaching their target. Consider six agents whose internal dynamics are given by
\begin{align*}
    \dot{x}_j= v_j \cos \delta_j,\;\;\;
    \dot{y}_j= v_j \cos \delta_j,\;\;\;
    \dot{\delta}_j = \omega_j.
\end{align*}
Here, the control inputs are $v_j$ and $\omega_j$, which are the linear and angular velocities of agent $j$, respectively.

We use the integrator dynamics defined in \eqref{eq:multiagent_controller} where $\bx_j = [x_j, \;y_j]$ and $\bu_j = [u_{j_x}, \;u_{j_y}]$. The density function for each agent is defined using \eqref{density_fun1}, where every other agent is modeled as a disk. Then, $v_j$ and $\omega_j$ can be obtained as follows \cite{moyalan2024synthesizing}:
\begin{align*}
    v_j = \sqrt{u_{j_x}^2+u_{j_y}^2},\;\;\;
    \omega_j = \dot{\Tilde{\delta}}_j - K(\delta_j-\Tilde{\delta}_j),
\end{align*}
where $K>0$ is a positive gain and $\Tilde{\delta} = \tan^{-1} \left(\frac{u_{j_y}}{u_{j_x}} \right)$. The control $\omega$ is designed such that the error $\left(\delta - \Tilde{\delta}\right)$ tends to zero asymptotically. This can be shown using a Lyapunov function as $V = \frac{1}{2}\left(\delta-\Tilde{\delta}\right)^2$.

\begin{figure*}
    \centering
    \includegraphics[width =1\linewidth]{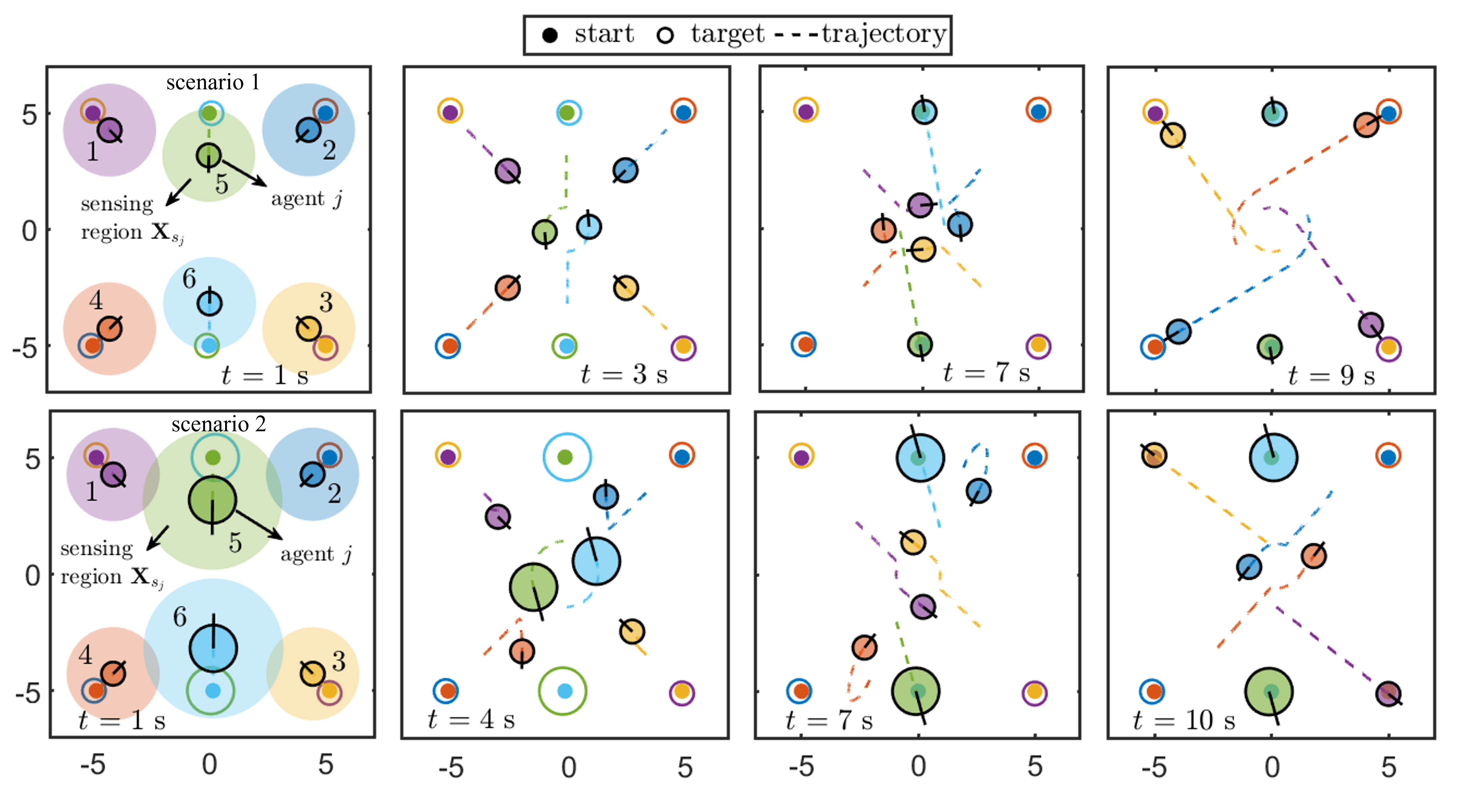}
    \caption{\textbf{Multi-agent Collision Avoidance:} Snapshots of interactions between six agents navigating an intersection while avoiding gridlock. Scenario 1 (top row) shows six identical agents, and Scenario 2 (bottom row) shows agents 5 and 6 having twice the radius as the other agents.}
    \label{fig:multiagent_unicycle_scenario}
\end{figure*}

Fig. \ref{fig:multiagent_unicycle_scenario} presents snapshots of six agents navigating an intersection and reaching their targets. Each agent is depicted as a solid circle, with a larger, transparent circle representing their sensing region. For clarity, we show the sensing region of each agent only in the first timestamp. Each agent's start and target positions are marked by a solid point and a hollow circle (with corresponding colors), respectively. The trajectories are shown as dashed lines, and the heading angles are indicated by solid lines extending from the agents. 

In scenario 1 (top row), all agents are modeled as disks with a radius $r_j = 0.5$ and a circular sensing radius of $s_j = 2$. Around $t = 3$ s, agents 5 and 6 resolve a conflict by executing circular maneuvers. The other agents initially slow down and resolve conflict by performing circular maneuvers around $t = 7$ s before converging on their targets after $t  = 9$s. In scenario 2 (bottom row), agents 5 and 6 have a radius of $r_5 = 1$ and $r_6=1$, respectively. The corresponding sensing radius is set to $s_5 =3$ and $s_6=3$, respectively. Agents 1 and 2 slow down initially and resolve conflicts around $t=7$ s. The larger agents (agents 5 and 6) push the smaller agents (agents 2 and 4) away from their shortest path trajectory (see timestamp $t=4$ s) before converging to the target. Eventually, agents 2 and 3 perform a course correction maneuver around $t=7$ s and resolve their conflicts around $t=10$ s before converging to their respective targets.
{\color{black}
\begin{remark}
    Since the convergence of each agent is with respect to almost everywhere sense, as defined by Definition \ref{def_aeuniformstable}, there is still a possibility of gridlock. This arises when the initial condition of agent $j$ lies in its zero-measure set. Obtaining theoretical guarantees for avoiding gridlock is left for future work.
\end{remark}
}

\subsubsection{Comparison with Social  Force Model} 
This section compares the simulation results for multi-agent collision avoidance obtained using our proposed approach with the social force model (SFM), which is a popular model for simulating collective human behavior \cite{helbing1995social}. For this comparison, we set up a scenario with four agents where each agent has to swap positions with the opposite agent. The agents are modeled as a disk of radius $r_j=0.75$ using double integrator dynamics $\dot{\bx}_j = \bv_j,\; \dot{\bv}_j = \bu_j$, where $\bx_j\in \mR^2$ and $\bv_j\in \mR^2$. First, we use the SFM to define the control for this system. This force-based method is used to represent human motion as a sum of desired forces, $f_{d_j}$ and repulsive forces, $f_{r_j}$ (as described in \cite{helbing2000simulating}).
\begin{equation}
    \bu_j = \bu_j^{SFM} := f_{d_j} + f_{r_j}.
\end{equation}
Next, we design a density-based controller for double integrator dynamics, using backstepping, as follows:
\begin{equation}
     \bu_j^{\rho}(t,\bx_j) = \frac{d}{dt} \left( \beta \nabla \rho_j(t,{\bx}_j) \right) - K\left(\bv_j - \beta \nabla \rho_j(t,\bx_j) \right),
\end{equation}
where $K>0$. The stability can be verified using the Lyapunov function $V = \frac{1}{2}\left(\bv_j - \beta \nabla \rho(t,\bx_j)\right)^2$. 

\begin{figure*}[ht]
    \centering
    \includegraphics[width=1\linewidth]{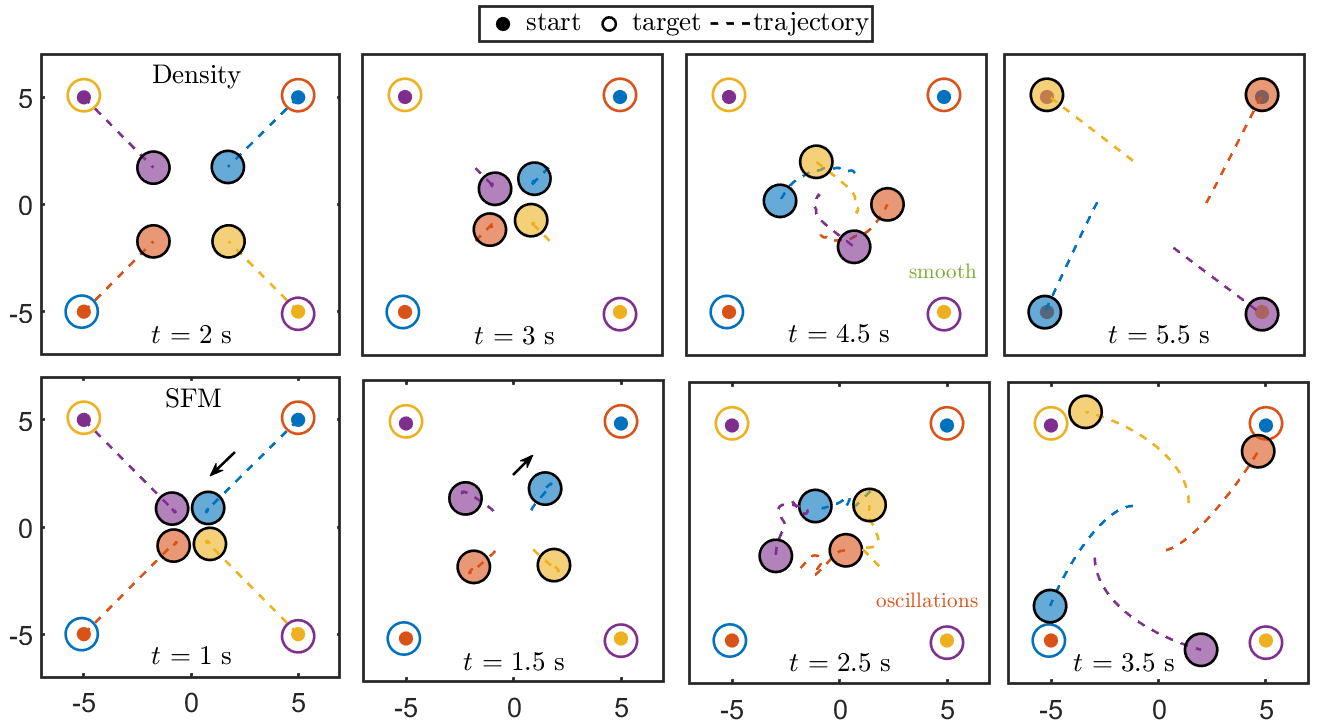}
    \caption{\textbf{Comparison with social force model:} The agents using the density-based controller resolve conflicts and converge to the target smoothly (top row), while the SFM-based controller can lead to oscillations (bottom row).}
    \label{fig:density_vs_SFM}
\end{figure*}

Fig. \ref{fig:density_vs_SFM} shows snapshots of the interaction between the four agents (with unit mass and radius of 0.75) as they avoid conflict and converge to the target. In the top row, each agent uses the density-based controller $\bu_j^\rho$. We construct the density function $\rho_j(t,\bx_j)$ as defined in \eqref{eq:mutliagent_density}. For each agent, we use $r_j=0.75$, $s_j=2$ in the construction of $\Phi_j(t,\bx_j))$ and define $V_1^j = \|\bx_j-\bz_j\|^2$. Further, we use $\alpha=0.2$ and $\beta=20$, $K = 1$. The agents approach the conflict region at around $t=3$ s and execute a circular maneuver to resolve the conflict at $t=4.5$ s before converging to the target at $t=5.5$ s. The bottom row shows the multi-agent interaction when each agent executes the SFM-based controller $\bu_j^{SFM}$. We model each agent as a circular disk with $r_j=0.75$, $r_k=0.75$, and a sensing distance of $d_H=2$. Further, we use $A_j = 2000$, $B_j=0.08$, $\kappa_1=1.2 \times 10^5$ and $\kappa_2=2.4\times 10^5$.

While resolving conflicts, the SFM-based controller leads to oscillations (see timestamp $t = $ 2.5 s, bottom row) since the agents behave like particles. In contrast, the density-based controller results in smoother trajectories. Further, to the best of our knowledge, the SFM model does not provide any theoretical guarantees for ensuring safety and convergence.

\subsection{Robotic Arm Trajectory Tracking}
In this section, we extend the density-based controller proposed in Theorem \ref{theorem_main2} to fully actuated robotic systems whose dynamics can be expressed using the Euler-Lagrange equations. Consider an unconstrained system with $\mathbf{q}$ and $\dot{\mathbf{q}}$  being the position and velocity states, respectively. Let $\bQ$ be the configuration manifold of the robot and $\mathbf{q} \in \bQ$. For a two-link planar robotic arm, $\mathbf{q}\in \bS^1 \times \bS^1$ represents the angle position of each link. Let $\bM(\mathbf{q})$ be the inertia matrix and $\bH(\mathbf{q},\dot{\mathbf{q}})$ represent the Coriolis and gravity effects on the system. The dynamics of the system can defined as follows 
\begin{equation}
   \bM(\mathbf{q})\ddot{\mathbf{q}} +\bH(\mathbf{q},\dot{\mathbf{q}}) = \bu_\rho(t,\mathbf{q},\dot{\mathbf{q}}).
   \label{eq:eulerLagragnge}
\end{equation}
{\color{black} The general approach involves a hierarchical structure that separates navigation into two stages: motion planning and control. In motion planning, the task is to determine a collision-free path or trajectory through the feasible configuration space, starting from an initial state and ending at a desired goal. This trajectory serves as the input for the control stage, where several existing methods, such as model predictive control, can be used to design appropriate control actions to follow the planned path, ensuring dynamic feasibility.
}

In the following, we outline how the proposed density-based approach can be used as a motion plan for such systems. We first construct a time-varying density function in the configuration space $\rho_{o_q}$ with $L$ obstacles as follows:
\begin{align}
    \rho_{o_q}(t,\mathbf{q}) &= \frac{\Pi_{k=1}^L \Psi(\mathbf{q})}{\left(V_q(t,\mathbf{q})+\kappa\right)^\alpha}, \label{eq:density_joint}
\end{align}
where $V_{q}(t,\mathbf{q})$ is a distance function that encodes the target trajectory and $\Psi_k(\mathbf{q})$ are inverse bump functions used to represent the obstacles in the configuration space. Next, we generate a safe path that avoids static obstacles while tracking a time-varying target, using the following system
\begin{equation*}
    \dot{\bq} = \nabla \rho_{o_q}(t,\bq). 
\end{equation*}
The solution $\bq(t)$ serves as the safe motion plan in designing safe navigation frameworks. This approach has been successfully employed for the safe navigation of quadruped robots and ground vehicles in \cite{moyalan2023off,sriram2023ICC, moyalan2024synthesizing}.

{\color{black}
Next, we use a two-link robotic arm to track a time-varying target while avoiding static unsafe sets. The system's dynamics can be given by \eqref{eq:eulerLagragnge}. The mass and length of each link are set to unity. We assume that the system is fully actuated and the obstacles are present only in the position states. 

To construct the density function, the obstacles (circular with a radius of 0.2 units) are mapped to configuration space and approximated using inverse bump functions. Further, we use the following distance function 
\begin{align*}
    V_q(t,\mathbf{q}) = \Big(1-\cos(\bar{q}_1)\Big)^2 + \Big(1-\cos(\bar{q}_2) \Big)^2,
\end{align*} 
where $\bar{q}_i=q_i-q_{i_T}(t)$, and construct the density function using \eqref{eq:density_joint}. Here $[q_{1_T}(t),\; q_{2_T}(t)]$ represents the target trajectory. The objective is to track the trajectory of a time-varying target (in task space) given by $\mathbf{x}_T(t) = [0.5+\sin(t), \; -0.6-\cos(t)]$.

First, we design a safe path for this system that avoids all unsafe sets while tracking a time-varying target. Specifically, a safe motion plan is obtained as a solution to the system $\dot{\hat{\bq}}=\beta \nabla \rho_{o_q}(t,\hat{\bq})$. Next, we define a density-based inverse dynamics controller given by
\begin{align}
    \bu_\rho = \bM(\bq) \biggl( \ddot{\bq}_d(t) - \bK_p \be(t)  - \bK_v \dot{\be}(t)\biggr) + \bH(\bq,\dot{\bq}), \label{eq:inverse_dyn_density}
\end{align}
where $\be(t):= \bq(t)-\bq_d(t)$ and $\dot{\be}(t):= \dot{\bq}(t)-\dot{\bq}_d(t)$, $\bq_d(t)$ is the desired reference trajectory which is obtained from the motion plan. $\bK_p$ and $\bK_v$ are constant gain matrices. Using the controller defined in \eqref{eq:inverse_dyn_density}, the system dynamics defined in \eqref{eq:eulerLagragnge} reduces to 
\begin{equation}
   \ddot{\be} + \bK_v \dot{\be} + \bK_p \be = 0. \nonumber
\end{equation}
Hence, by choosing $\bK_p>0$ and $\bK_v>0$, it can be easily verified that the above system is asymptotically stable, i.e., the tracking errors $\be$ and $\dot{\be}$ go to zero asymptotically.
\begin{remark}
    Note that the system's safety is not guaranteed if the system is only asymptotically stable. However, with large enough gains and a suitable Lyapunov function, it can be shown that the system is exponentially stable, i.e., the error goes to zero in a finite time. 
\end{remark}

Fig. \ref{fig:time_varying_goal} shows snapshots of the system tracking a circular target trajectory while avoiding obstacles. We use $\beta=10$ to obtain the motion plan and set $\bK_p=1$ and $\bK_v=10$ for the inverse dynamics controller. The system starts tracking the target at $t=1$ s while avoiding the obstacle between $t=2$ s and $t=3$ s before converging to the target again at $t=4$ s. 

\begin{figure}
    \centering
    \includegraphics[width =\linewidth]{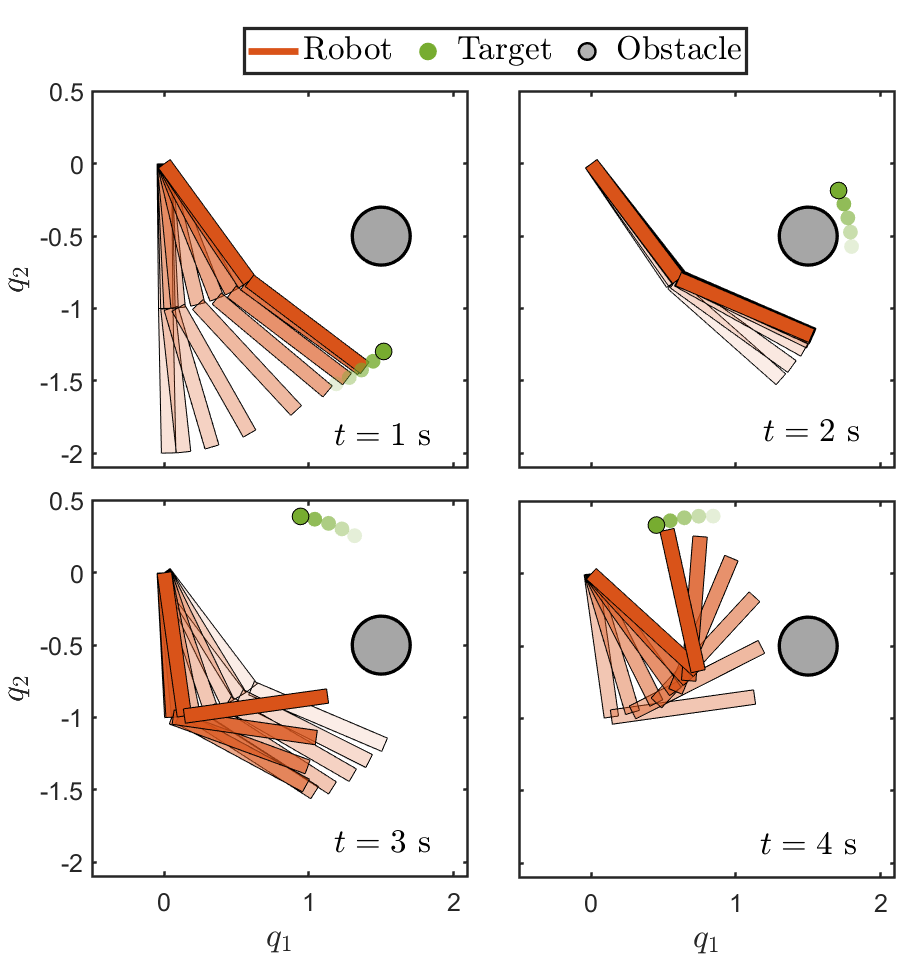}
    \caption{\textbf{Robotic Arm Trajectory Tracking with safety:} Snapshots of the robot (red) tracking a time-varying target $\mathbf{x}_T(t)$ (green) while avoiding the unsafe set $\bX_u$.}
    \label{fig:time_varying_goal}
\end{figure}
}

\section{CONCLUSIONS}\label{sec:conclusions}
This paper presents a novel framework for safe navigation in dynamic environments. We propose an analytical method for constructing density functions tailored to two critical scenarios: (i) static targets in the presence of time-varying obstacles, and (ii) time-varying target trajectories in the presence of static obstacles. At the core of our approach is a density-based feedback controller that ensures both safety and almost-everywhere convergence of system trajectories. The effectiveness of the proposed method is demonstrated through two representative applications: collision avoidance in multi-agent systems and safe trajectory tracking for a two-link planar robotic arm. These results highlight the practical applicability of our framework to real-world robotic control problems, particularly in environments that demand dynamic adaptability and safety guarantees.

\section{Appendix} \label{sec:appendix}
\textbf{Proof of Lemma \ref{lemma:ineq}:}
In this proof, we illustrate how to select values of $\alpha$ and $\beta$ to satisfy the inequalities \eqref{eq:x_sensing} and \eqref{eq:x_not_sensing} to satisfy Lemma \ref{lemma:ineq}. We note that any sufficiently small positive value of $\theta$ will satisfy Lemma \ref{lemma:ineq}. The values of $\alpha$, $\beta$ and $\theta$ used in practice are provided in Remark \ref{remark_tuning}.

{\bf Range of} $\alpha$: We will find the range of $\alpha$ which satisfies the inequalities defined in \eqref{eq:subs4} and \eqref{eq:subs3_pre}. 

\underline{Inside the sensing region}, i.e., when $\bx \in \bX_{s_k}$, we first illustrate how we arrive at \eqref{eq:subs4}. Let us recall the inequality defined in \eqref{divergence_final_term}, and rewrite it compactly as
\begin{align*}
    &\nabla\cdot(\bk\rho_o) =\; \frac{\alpha\beta}{V^\alpha V_1^\alpha} 
    \sum_{j=1}^n \left(q_1 + q_2 + q_3 + q_4 \right),\\
    \textrm{where,}\\
    &q_1 := 
    \left( \frac{\alpha}{VV_1} + \frac{\alpha+1}{V_1^2} \right)\Psi^2 
    \left( \frac{\partial V}{\partial x_j} \right)^2, \\
    &q_2 := -\left( \frac{3}{V_1} + \frac{1}{V} \right)\Psi 
    \frac{\partial V}{\partial x_j} \frac{\partial \Psi}{\partial x_j},\\ 
    &q_3 := -\frac{\Psi^2}{V_1} \frac{\partial^2 V}{\partial x_j^2}, \\
    &q_4 := \frac{1}{\alpha} \left( 
    \Psi \frac{\partial^2 \Psi}{\partial x_j^2} + 
    \left( \frac{\partial \Psi}{\partial x_j} \right)^2 \right).
\end{align*}
Using Assumption~\ref{assume_main}.4 and \ref{assume_main}.5, we get the following bounds for each term in \eqref{divergence_final_term}:
\begin{align}
    &q_1\ge (2\alpha+1) \left( \bar{d}_V\|\bx\|^{2} + \kappa \right)^{-2} \theta^2 (\underline{d}_{V_x})^2 \|\bx\|^2 
    \label{ineq:1_pre}, \\
    &q_2 \ge -4 (\underline{d}_{V})^{-1} \bar{d}_{V_x} \bar{c}_{\Psi_x} 
    \|\bx\|^{-1}, 
    \label{ineq:2_pre} \\
    &q_3 \ge -\kappa^{-1} \bar{d}_{V_{x^2}} 
    \label{ineq:3_pre}, \\
    &q_4 \ge - \frac{1}{\alpha}\bar{c}_{\Psi_{x^2}} 
    \label{ineq:4_pre}.
\end{align}
Further, using Assumption~\ref{assume_main}.2 and ~\ref{assume_main}.5, we can rewrite \eqref{ineq:1_pre}-\eqref{ineq:2_pre} as follows:
\begin{align}
    &q_1 \ge (2\alpha+1) \left( \bar{d}_V(\bar{c}_x)^2 + \kappa \right)^{-2}(\underline{d}_{V_x}\,\bar{c}_x\,\theta)^2, 
    \label{ineq:1} \\
    &q_2 \ge -4 (\underline{d}_{V})^{-1} \bar{d}_{V_x} 
    \bar{c}_{\Psi_x} (\underline{c}_x)^{-1}. 
    \label{ineq:2}
\end{align}
Therefore, utilizing \eqref{ineq:3_pre}-\eqref{ineq:2}, we get the following lower bound defined in \eqref{eq:subs4} for all $\bx \in \bX_{s_k}$
\begin{align}
    &\nabla \cdot (\bk \rho) 
    \ge \frac{\alpha \beta n}{V^\alpha V_1^\alpha} \bigg( 
    (2\alpha + 1) \left( \bar{d}_V (\bar{c}_x)^2 + \kappa \right)^{-2}  (\underline{d}_{V_x}\,\bar{c}_x\,\theta)^2 \nonumber \\
    &\quad - 4 (\underline{d}_V)^{-1} \bar{d}_{V_x} \bar{c}_{\Psi_x} 
    (\underline{c}_x)^{-1} 
    - \kappa^{-1} \bar{d}_{V_{x^2}} 
    - \alpha^{-1}\,\bar{c}_{\Psi_{x^2}} 
    \bigg). \nonumber
\end{align}
Next, we can write \eqref{eq:subs4} compactly as follows:
\begin{align}
    &\nabla \cdot (\bk \rho) \geq \frac{\beta n}{V^\alpha V_1^{\alpha}}\left(p_1\alpha^2 + p_2 \alpha - p_3\right), \label{eq:subs33} \\
    \textrm{where,} \nonumber\\
    \quad \quad p_1 := &\; 2(\bar{d}_V(\bar{c}_x)^{2}+ \kappa)^{-2} (\underline{d}_{V_x}\,\bar{c}_x\,\theta)^2, \nonumber\\
    p_2 :=&\; (\bar{d}_V(\bar{c}_x)^{2}+ \kappa)^{-2} (\underline{d}_{V_x}\,\bar{c}_x\,\theta)^2 \nonumber\\
    &- 4(\underline{d}_{V})^{-1}\bar{d}_{V_x}\bar{c}_{\Psi_x}(\bar{c}_x)^{-1}  - \kappa^{-1}\bar{d}_{V_{x^2}}, \nonumber\\
    p_3 := &\; \bar{c}_{\Psi_{x^2}}. \nonumber
\end{align}
We observe that for a sufficiently small value of $\theta$, the term inside the brackets in \eqref{eq:subs33} is a convex quadratic function, as its Hessian is given by $2p_1>0$.  Further, we note that $p_1>0$ and $p_3>0$. Although we cannot comment on the sign of $p_2$, we know that $p_2^2\geq0$. Consequently, the discriminant $p_2^2 + 4p_1p_3$ is strictly positive, and $\sqrt{p_2^2 + 4p_1p_3} > -p_2$. Therefore, the condition under which the lower bound in \eqref{eq:subs33} remains positive is given by the positive root of the quadratic
\begin{align}
    \alpha > \frac{-p_2+\sqrt{p_2^2+4p_1p_3}}{2p_1} > 0. \label{range2}
\end{align}
Note that for the case when $p_2=0$ (and hence $p_2^2=0$), the range of $\alpha$ reduces to $\alpha>p_3$.


\underline{Outside the sensing region}, i.e., when $\bx \notin \bX_{s_k}$, we note that $\bar{d}_V\|\bx\|^2+\kappa \le \bar{d}_{V_1}\|\bx\|^2$, where we can choose $\bar{d}_{V_1} \ge \bar{d}_V + \kappa\,\delta^{-1} $
Therefore, we can rewrite \eqref{eq:subs3_pre} as follows
\begin{align}
\nabla\cdot(\bk\rho)\ge &\frac{\alpha\beta n \|\bx\|^{-2}}{V^\alpha V_1^\alpha} \bigg((2\alpha+1)(\bar{d}_{V_1})^{-2}  (\underline{d}_{V_x}\,\theta)^2\nonumber \\
&-(\underline{d}_{V})^{-1}\bar{d}_{V_{x^2}}\bigg). \label{eq:subs3}
\end{align}
Next, we show that for a sufficiently small value of 
$\theta$, there exists a value of $\alpha$ such that the lower bound in \eqref{eq:subs3} remains positive. To guarantee this condition, we have the following condition on $\alpha$  
\begin{align}
    \alpha > \frac{(\underline{d}_{V})^{-1}\bar{d}_{V_{x^2}}}{2(\bar{d}_{V_1})^{-2}  (\underline{d}_{V_x}\,\theta)^2}-\frac{1}{2}.\label{range1}
\end{align} 
Therefore, from \eqref{range2} and \eqref{range1}, we obtain the following condition on the choice of $\alpha$:
\begin{align}
   \alpha > \max\left \{\frac{-p_2+\sqrt{p_2^2+4p_1p_3}}{2p_1}, \frac{(\underline{d}_{V})^{-1}\bar{d}_{V_{x^2}}}{2(\bar{d}_{V_1})^{-2}  (\underline{d}_{V_x}\,\theta)^2}-\frac{1}{2}\right \}. \label{alpha_choice}
\end{align} 
{\color{black}{\bf Range of} $\beta$: To find the range of $\beta$ values, lets first recall from \eqref{ineq:55} and \eqref{ineq:555}.
\begin{align*}
\left|\frac{\partial \rho_o}{\partial t}\right|&=\frac{1}{V_1^\alpha} \left| \frac{\partial \Psi}{\partial t} \right| \le \frac{1}{V_1^\alpha}c_{\psi_t} &\quad \forall \; \bx \in \bX_{s_k}, \\
\frac{\partial \rho_o}{\partial t} &= 0 &\quad  \forall \; \bx \notin \bX_{s_k}.
\end{align*}
\underline{Inside the sensing region}, i.e., when $\bx \in \bX_{s_k}$, we first rewrite \eqref{eq:subs4}, after factoring out $\beta$ and $V_1^\alpha$ as follows
\begin{align}
    \nabla\cdot(\bk(\bx,t)\rho_o)\geq \frac{\beta L_1}{V_1^\alpha} >0,\label{ineq:77}
\end{align}
where,
\begin{align}
L_1 &:= \alpha n (\bar{d}_V)^{-1} (\bar{c}_x)^{-2} \Big( (2\alpha{+}1)(\bar{d}_V (\bar{c}_x)^2{+}\kappa)^{-2} (\underline{d}_{V_x}\,\bar{c}_x\,\theta)^2 \nonumber \\
&\quad - 4\,(\underline{d}_V)^{-1} \bar{d}_{V_x} \bar{c}_{\Psi_x} (\bar{c}_x)^{-1} - \kappa^{-1} \bar{d}_{V_{x^2}} - \alpha^{-1} \bar{c}_{\Psi_{x^2}} \Big). \nonumber
\end{align}
Note that $L_1>0$ for the given choice of alpha in \eqref{alpha_choice}. Therefore, when $\bx \in \bX_{s_k}$, the following inequality should be satisfied for $a.e. \;\; (t,\bx)\in \mathbb{R} \times \bX_1$: 
$$\frac{1}{V_1^\alpha}(-c_{\Psi_t} + \beta L_1) > 0.$$
Hence, the lower bound on $\beta$ which satisfies this condition can be obtained as:  
\begin{align}
    \beta > \frac{c_{\Psi_t}}{L_1}. \label{beta_choice}
\end{align}

\underline{Outside the sensing region}, i.e., when $\bx \notin \bX_{s_k}$, Theorem \ref{theorem_main} can be satisfied $\forall \beta > 0$. Hence, \eqref{beta_choice} provides the overall range of $\beta$. 

\bibliographystyle{unsrt}  
\bibliography{references_tac}

\end{document}